\documentclass{article}

     \PassOptionsToPackage{numbers, compress}{natbib}
\usepackage[preprint]{neurips_2026}

\usepackage[utf8]{inputenc} % allow utf-8 input
\usepackage[T1]{fontenc}    % use 8-bit T1 fonts
\usepackage{url}            % simple URL typesetting
\usepackage{booktabs}       % professional-quality tables
\usepackage{amsfonts}       % blackboard math symbols
\usepackage{nicefrac}       % compact symbols for 1/2, etc.
\usepackage{microtype}      % microtypography
\usepackage{xcolor}         % colors

 \usepackage{amsmath}
 \usepackage{graphicx}
 \usepackage{wrapfig}
 \usepackage[most]{tcolorbox}
 \usepackage{booktabs}
 \usepackage{makecell}
 \usepackage{multirow}
\usepackage{colortbl}
\usepackage{amssymb}
\usepackage{arydshln}
\usepackage{titletoc}
\usepackage[dvipsnames]{xcolor}
\usepackage{tabularx}
\usepackage[noend]{algpseudocode}
\usepackage{algorithm}
\usepackage{listings}

\usepackage{threeparttable}
\title{Structuring Open-Ended NAS: Semi-Automated Design Knowledge Structuring with LLMs for Efficient Neural Architecture Search}

\author{%
   Yuiko Sakuma\textsuperscript{1} \quad
   Masakazu Yoshimura\textsuperscript{1} \quad
   Marcel Gr\"opl\textsuperscript{2}\quad\\
   \textbf{
   Zitang Sun\textsuperscript{1} \quad
   Junji Otsuka\textsuperscript{1} \quad
   Atsushi Irie\textsuperscript{1} \quad
   Takeshi Ohashi\textsuperscript{1} \quad
   }\\
    \textsuperscript{1}Sony Group Corporation, Tokyo, Japan \quad
    \textsuperscript{2}ETH Z\"urich \\
   \texttt{Yuiko.Sakuma@sony.com}
}
\begin{document}

\maketitle

\begin{abstract}
Current neural architecture search (NAS) methods are often limited by their predefined, restrictive search spaces. While recent large language model (LLM)-assisted NAS methods enable open-ended search spaces, they often suffer from inefficient exploration due to biased or low-quality design ideas. To address these issues, we propose to semi-automatically structure model design knowledge to guide the search process. Our approach first defines a high-level structural template of architectural attributes. An LLM then populates this template by analyzing papers, creating a rich and diverse search space that embodies this structured design knowledge. To efficiently explore this vast space, we introduce FairNAD, using a multi-type mutation that enables broad exploration through mutation with fair idea sampling, Pareto-aware mutation, LLM-driven iterative mutation, and a fine-grained feedback loop. We demonstrate the effectiveness of FairNAD in discovering high-performing architectures that yield 0.84, 2.17, and 2.35 points improvement on CIFAR-10, CIFAR-100, and ImageNet16-120, respectively, compared to current state-of-the-art methods.
\end{abstract}

\section{Introduction}

Designing high-performance neural network architectures remains a challenge, as manual architecture design heavily relies on an extensive trial-and-error process conducted by experts. This process is not only time-consuming but also confines the search to the scope of the designer's expertise, which limits the performance improvements of the resulting models. To address these issues, Neural Architecture Search (NAS) has been extensively investigated as an approach to automate the design process, aiming to reduce search costs while enabling a broader search \cite{liu2018darts,zoph2017nasrl}. However, many conventional NAS methods depend on a closed search space, predefined by human experts. This means the search is restricted to combinations of existing operators and blocks, fundamentally limiting the ability to generate innovative architectures.

In recent years, a new paradigm has emerged to overcome this limitation: leveraging Large Language Models (LLMs) for NAS. LLM-based NAS adopts an approach of directly generating the source code of a neural network model, either entirely or partially \cite{chen2023evoprompting, nasir2024llmatic, zheng2023can, yang2025nader}. This enables an ``open-ended'' search, removing the limitations imposed by the predefined search space in traditional methods. As a result, LLM-based NAS facilitates the discovery of novel operations, modules, and network topologies, potentially enabling the design of high-performing architectures. Moreover, recent studies \cite{yang2025nader} have shown that external design knowledge, such as architectural knowledge extracted from papers, can further enrich this open-ended search process.

However, open-endedness alone does not guarantee efficient exploration. The effectiveness of knowledge-guided open-ended NAS critically depends on how the underlying design knowledge is collected, organized, and sampled. In existing methods, design knowledge is directly obtained from LLM priors \cite{zheng2023can}, network architectures \cite{zhou2025design}, papers \cite{yang2025nader}, or web resources \cite{cheng2025language}, and is used in a relatively flat and unstructured manner. These methods retrieve and utilize design knowledge without considering its scope of application (e.g., whether it pertains to an operation, connectivity or the entire network) and its attribute coverage (e.g., neural operators may fall into categories such as convolutions, Transformers, multi-layer perceptrons, and pooling). As a result, knowledge with different levels of abstraction, quality, and applicability are mixed together, making it difficult to effectively retrieve and utilize useful knowledge. Consequently, the search space may include noisy, low-quality, or overly domain-specific knowledge, and may also inherit biases from LLM corpora and research trends. For example, if the knowledge pool is largely constructed from papers published during a period dominated by Transformer-based improvements, the resulting search process may overemphasize that direction and narrow the scope of exploration.

\begin{figure}
  \centering
  \includegraphics[width=0.95\linewidth]{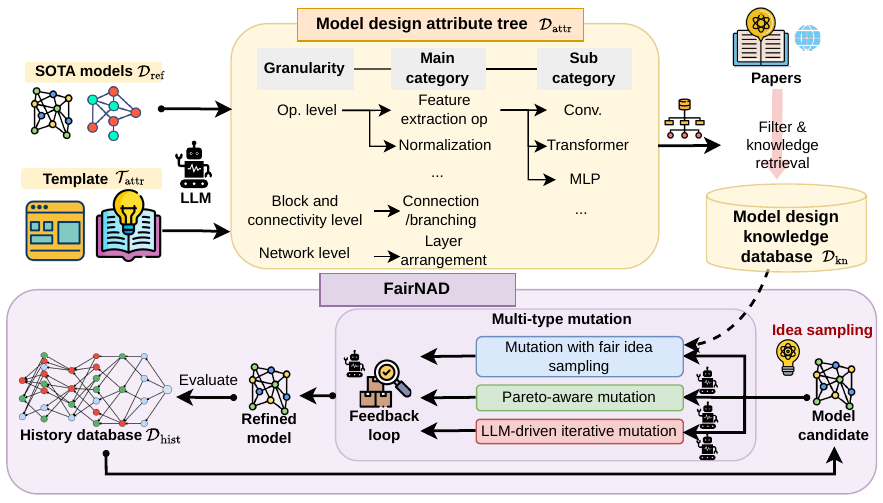}
  \caption{\textit{Overview of the proposed NAS.} (\textbf{Top}) The model design attribute tree is generated from state-of-the-art (SOTA) models and a structured template. This tree is used to extract high-quality, fine-grained model design knowledge. (\textbf{Bottom})  FairNAD, a LLM-driven framework, then searches for high-performing models using mutation with fair idea sampling, Pareto-aware mutation, LLM-driven iterative mutation, and feedback loop.}
    \label{fig:overview}
\end{figure}

In this work, we address the inefficiency of open-ended NAS by proposing to structure the design knowledge for open-ended NAS. In contrast to prior open-ended methods such as \cite{yang2025nader}, which draws design knowledge from an unstructured knowledge pool, we curate the knowledge using a structured model design attribute template. This structured representation enables filtering of overly domain-specific knowledge, collection of higher-quality design ideas into a model design knowledge database, and mitigation of biases inherited from LLM priors and research trends. Further, in a vast open-ended search space, unstructured and performance-driven search strategies remain inefficient. In particular, strategies that sample design ideas from a flat knowledge pool, select parent architectures mainly according to validation performance, and discard candidates that violate HW constraints may suffer from three limitations: (1) they overlook diverse model candidates, particularly failing to select compact yet high-performing models for their size; (2) they lack intensive exploration of promising models; and (3) diversity is reduced by discarding models that exceed hardware (HW) constraints. Thus, building on the structured knowledge space, we propose FairNAD, a framework which employs a multi-stage search process for efficient and effective open-ended neural architecture design (NAD). First, FairNAD introduces the \textit{mutation with fair idea sampling} stage which enables a category-balanced and uniform exploration of a vast design space by retrieving design ideas based on the structured model knowledge pool. In particular, the idea is sampled according to its attributes, referred to as fair idea sampling.

Second, the \textit{Pareto-aware mutation} stage performs Pareto-aware sampling under HW constraints, together with upscaling and hyperparameter adjustment, so that compact yet promising models are not overlooked during the evolution. Third, the \textit{LLM-driven iterative mutation} stage intensively refines the most promising models to accelerate convergence. Furthermore, the \textit{feedback loop} is incorporated to downsize any models that exceed HW constraints, thereby encouraging the diversity of candidate models by retaining models that would otherwise be discarded.

We demonstrate the effectiveness of FairNAD in designing high-performing architectures by exploring the open-ended search space efficiently and effectively. The experiments on NAS-Bench-201 achieves state-of-the-art performance over previous methods. Our contributions are summarized as follows:
\begin{itemize}
    \item We introduce a semi-automated model design attribute structuring method that organizes design knowledge into a hierarchical attribute tree. Based on this structured representation, we curate and organize diverse model design ideas from papers.
    \item We introduce FairNAD, using a multi-type mutation designed to efficiently explore an open-ended search space. First, the \textit{mutation with fair idea sampling} stage uniformly explores a wide range of model design ideas based on their design attribute. Second, the \textit{Pareto-aware mutation} stage scales up compact yet high-performing models. Third, the \textit{LLM-driven iterative mutation} stage intensively refines the most promising models. This multi-stage approach enhances both the diversity of candidate models and search efficiency. Furthermore, each model generation process incorporates a \textit{feedback loop}. This loop verifies and refines generated models. In particular, models exceeding the HW constraints are scaled down which further promotes architectural diversity.
\end{itemize}

\section{Related Works}

\subsection{Neural Architecture Search (NAS)}
Neural Architecture Search (NAS) aims to automatically explore optimal neural network architectures. Early NAS work demonstrated that reinforcement-learning controllers can generate competitive architectures \cite{pham2018enas, zoph2017nasrl,  zoph2018nasnet}, and that evolutionary strategies can discover strong image classifiers at scale \cite{liu2018pnas, so2019evolved, suganuma2017genetic}. Supernet-based NAS typically adopts a weight-sharing strategy to avoid training each subnet from scratch \cite{cai2018proxylessnas, cai2020once, chu2021fairnas, liu2018darts, stamoulis2019single, wu2019fbnet, yu2020bignas}. Prior work has also extended supernet-based NAS to Transformer architectures \cite{chen2021autoformer, chen2021searching}. Recently, GNN-based methods have been proposed for efficient optimization \cite{mills2023aio, mills2024building, salameh2023autogo}. However, these methods are constrained by their search spaces, which limit their ability to discover novel architectures.

\subsection{LLM-Based NAS}
Recent works propose to use LLMs in evolutionary-guided NAS by directly prompting them to modify the architecture. Initial attempts \cite{chen2023evoprompting, li2025collm, nasir2024llmatic, zheng2023can,rahman2024lemo} rely on predefined prompts or the LLM's internal knowledge, resulting in a narrow search space or an inherent dependence on the LLM's pretrained knowledge. To broaden this, some methods \cite{cheng2025language, yang2025nader, zheng2023can, zhou2025design}, incorporate external sources for exploration. For example, \cite{zhou2025design} transfers the design principles of a predefined set of SOTA architectures to initialize the search process. Going further, \cite{cheng2025language, yang2025nader} mine current papers or web sources, achieving broader architectural diversity. However, this grounding in recent publications introduces a bias toward research trends, potentially limiting discoverable architectures.

\section{Problem Statement}

This paper addresses search efficiency in open-ended NAS, a paradigm that moves beyond predefined, hand-crafted search spaces. Our open-ended search is enabled by directly generating source code of the architectures (e.g., in PyTorch) using an LLM. Unlike conventional hand-crafted NAS, which composes architectures from a predefined set of blocks (e.g., 3x3/5x5 convolutions, Transformer layers) and adheres to fixed topologies (e.g., ResNet-based), an LLM-based method enables the design of new blocks and modifications to the fundamental dynamics. We term this exploration process within open-ended NAS as Neural Architecture Design (NAD). 
Our approach contrasts with prior LLM-based code generation methods that are ideally NAD, such as GENIUS \cite{zheng2023can} and LLMatic \cite{nasir2024llmatic}. These operate on a limited set of prompts and thus represent a more constrained problem setting. Our method, following prior work such as NADER \cite{yang2025nader}, constructs the search space from diverse external sources like papers, enabling code generation from a wide variety of prompts.

The scope of this research focuses on discovering backbones for vision models. We do not address the design of the neck and head architectures for downstream tasks. Architecture design is defined as a search problem to find a valid model of optimal architecture $a^{\ast}$ that maximizes a metric subject to HW constraints such as floating-point operations per second (FLOPs) or parameter size. A valid model is defined as a syntactically correct, executable (e.g., can be initialized, differentiable, and forward and backward defined), and multi-layer PyTorch model code.
%The optimization can be formulated as:
%\begin{equation}
%    a^\ast = \mathrm{argmax}_{a\in \mathcal{A}} \{ F^{\mathrm{fit}}(a)\}, \quad \text{s.t.} \quad \mathrm{budget}_i(a) \leq \tau_i, \quad \forall i \in \{1, \dots, N\}
%\end{equation}
%where $\mathcal{A}$ is the set of architecture codes in open-ended search space and $\tau_i$ is the budget of each HW constraint.

%The input to the proposed NAD framework is the base architecture code $a^{\mathrm{base}}$ which represents a commonly-used high-performing architectures such as ResNet \cite{he2016deep}. The output of the framework is the optimized architecture $a^{\ast}$. The search process can be achieved by a evolutionary search process \cite{real2017large}, which repeatedly rewrite the program code using a LLM.

\section{Method}

We propose an LLM-based NAS method as illustrated in Fig. \ref{fig:overview}. The overall procedure is as follows.
First, the source code of the existing state-of-the-art models is analyzed using an LLM to construct a model design attribute tree ($\mathcal{D}_{\mathrm{attr}}$). This process identifies the attributes that describe how existing architectures are composed and what elements they use, and enumerates potential operations for improvement.
Second, papers are analyzed using an LLM combined with $\mathcal{D}_{\mathrm{attr}}$ to generate diverse model design ideas. At the same time, papers that are irrelevant to architectural design are filtered out, resulting in the construction of a curated and structured model design knowledge database ($\mathcal{D}_{\mathrm{kn}}$).
Third, in the evolutionary search phase, a parent model that has been previously generated and evaluated is sampled from the evolution tree, while modification ideas are sampled from $\mathcal{D}_{\mathrm{kn}}$. Based on these ideas, multi-type mutations are performed.

In this paper, there are two main proposals over previous work: (1) enhancing the idea collection phase through the semi-automatic structuring of model design knowledge and (2) introducing a FairNAD framework that integrates (a) mutation with fair idea sampling, (b) Pareto-aware mutation, (c) an LLM-driven iterative mutation, and (d) combining the search process with a feedback loop.

\subsection{Semi-automated Model Design Attribute Structuring} \label{sec:4.1_semi_automated_attr_structuring}

\begin{wrapfigure}{r}{0.35\textwidth}
    \centering
    \begin{minipage}{\linewidth}
      \centering
      \includegraphics[width=1.0\linewidth]{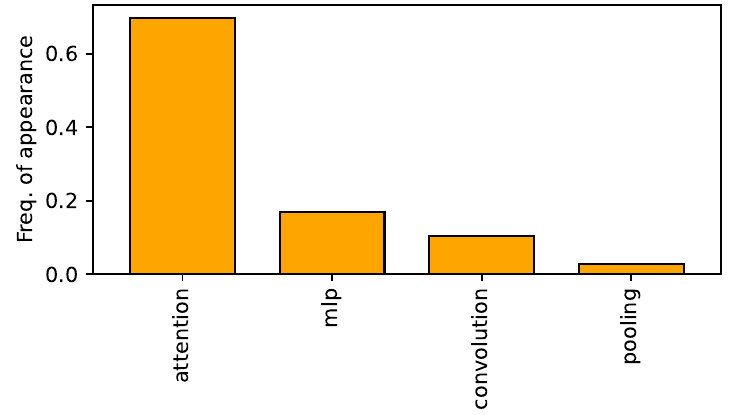} \\ 
      \includegraphics[width=1.0\linewidth]{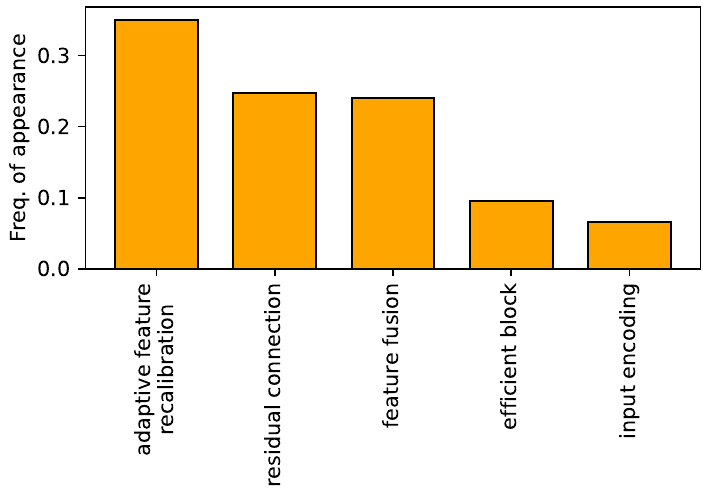}
    \end{minipage}
    \caption{Example of frequency of model design ideas for (\textbf{top}) feature extracting operations and (\textbf{bottom}) block and connectivity.}
    \label{fig:category example}
\end{wrapfigure}
Extracting model design knowledge from external sources, such as papers, by simply prompting an LLM with a general query like ``extract model design ideas from this paper'' can lead to outputs heavily biased by the LLM's internal knowledge and research trends.
To illustrate this, we analyzed the frequency of specific keywords in the attributes of ideas within $\mathcal{D}_{\mathrm{kn}}$ constructed from the sources  used in the prior work \cite{yang2025nader}. These attributes were tagged by the method described later in this section. As shown in Figure \ref{fig:category example}(\textbf{top}), approximately 70\% of the ideas for feature extraction operations are related to attention. Similarly, Figure \ref{fig:category example}(\textbf{bottom}) indicates that most ideas about blocks and connectivity are related to adaptive feature recalibration. Furthermore, as illustrated in Figure \ref{fig:NAS_comparison} (\textbf{(middle)}), some ideas are irrelevant to architecture design (e.g., including terms like ``loss'') or overly task- or domain-specific (e.g., including terms like ``3D'').

To address this challenge and facilitate the extraction of broad and comprehensive knowledge relevant to architecture design, we employ a structured model design attribute tree, $\mathcal{D}_{\mathrm{attr}}$, as an example when querying the LLM. The creation of this tree presents a challenge; a purely hand-crafted design would be constrained by the engineer's knowledge, whereas a fully automated generation using an LLM would be biased. Therefore, we propose a semi-automated approach. This method involves analyzing commonly used performant models as a reference with a predefined attribute template to systematically list model design attributes, striking a balance between manual design and full automation.

As illustrated in Figure \ref{fig:overview} (\textbf{top}), the proposed model design attribute generator takes two inputs: an attribute template ($\mathcal{T}_{\mathrm{attr}}$) and a set of reference model codes (${\mathcal{D}}_{\mathrm{ref}}$). The output is a model design attribute tree (${\mathcal{D}}_{\mathrm{attr}}$). We formulate a three-level tree structure within $\mathcal{T}_{\mathrm{attr}}$. This hierarchy consists of granularity (i.e., operation, block and connectivity, network), main attributes (i.e., categories specific to each granularity, such as feature extraction operation at the operation level), and sub-attributes (i.e., finer-grained categories within each main attribute, such as grouped convolution). The top two levels (e.g., granularity and main category) were predefined based on expert knowledge, while the sub-attributes were generated by prompting an LLM. For each reference model $a^{\mathrm{ref}}\in {\mathcal{D}}_{\mathrm{ref}}$, the LLM is queried with $\mathcal{T}_{\mathrm{attr}}$ to extract the corresponding sub-attributes. $\mathcal{T}_{\mathrm{attr}}$ was designed for different targets, performance and efficiency improvements. 
%Full details of the prompts can be found in Appendix \ref{sec:F1.1_attribute_template}.

\subsection{Knowledge Extraction Using Model Design Attribute Tree}

\begin{figure}[t]
  \centering
  \includegraphics[width=0.9\linewidth]{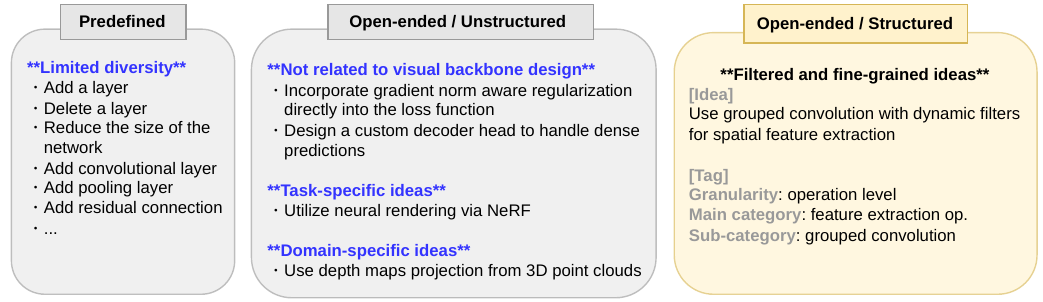}
  \caption{The difference of the model design ideas between (\textbf{left}) predefined \cite{nasir2024llmatic}, (\textbf{middle}) open-ended and unstructured \cite{yang2025nader}, and (\textbf{right}) open-ended and structured methods.}
  \label{fig:NAS_comparison}
\end{figure}

To build a high-quality, broad, and diverse knowledge base, we utilize our attribute tree to distill design ideas from papers. The process is designed to filter overly domain- or task-specific and irrelevant ideas related to architectural design and to encourage the extraction of fine-grained and diverse ideas from papers. This module takes a paper pool and $\mathcal{D}_{\mathrm{attr}}$ as input, and constructs a model design knowledge database, $\mathcal{D}_{\mathrm{kn}}$.
The procedure begins with a filtering stage. After scraping papers to structure the paper pool, the LLM is queried to classify each abstract against $\mathcal{D}_{\mathrm{attr}}$. Papers that fail to align with any attribute are discarded. In the subsequent extraction stage, the full text of the filtered papers is analyzed. The LLM is queried to read the method section if the abstract refers to any $\mathcal{D}_{\mathrm{attr}}$ element. The LLM is then queried to extract the associated model design ideas that can inspire architecture design. Potentially, multiple ideas are extracted from a single paper. Each idea is tagged with the corresponding element at each level of the hierarchy of $\mathcal{D}_{\mathrm{attr}}$ (i.e., granularity, main category and sub-category). The resulting ideas are then stored in $\mathcal{D}_{\mathrm{kn}}$. 

The examples of model design ideas from different methods are illustrated in Figure \ref{fig:NAS_comparison}. While predefined methods \cite{nasir2024llmatic} rely on a limited set of design concepts, open-ended methods \cite{yang2025nader} explore a diverse range of ideas, offering the potential to explore a wider variety of architectures. However, a challenge with these open-ended methods is that some generated ideas can be excessively domain- or task-specific, or even irrelevant to visual backbone design. Our proposed method, which is both open-ended and structured, addresses this limitation by systematically curating these ideas. Furthermore, it tags each idea with specific attributes that are utilized later for effective sampling. 
%For a detailed description of the prompts and more examples of generated ideas, please refer to the Appendix.

\subsection{FairNAD}

\begin{figure}
  \centering
  \includegraphics[width=0.9\linewidth]{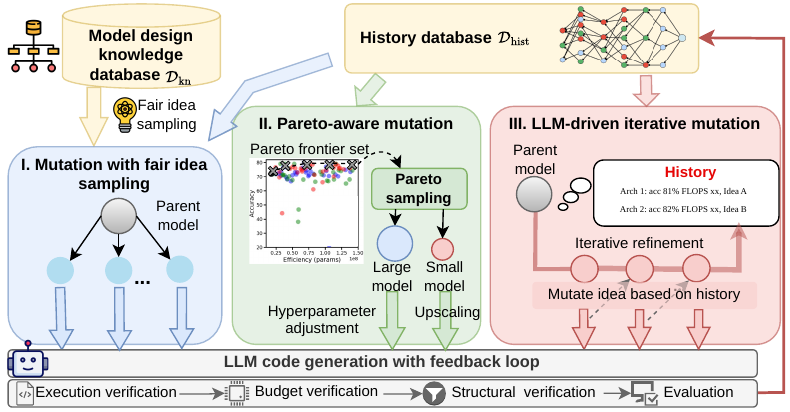}
  \caption{\textit{FairNAD employs a multi-type mutation to balance exploration and exploitation.} (\textbf{I}) Model design ideas are uniformly sampled according to its attributes. (\textbf{II}) To explore models on the Pareto frontier, small models are scaled up, while large models undergo hyperparameter tuning. (\textbf{III}) An LLM agent then iteratively refines high-performing ideas and candidate models.}
  \label{fig:fairNAD}
\end{figure}

In open-ended NAS, simply mutating models using uniformly sampled ideas is inefficient for exploring the vast search space. To address this inefficiency, we propose FairNAD, using a multi-type mutation as illustrated in Figure \ref{fig:fairNAD}. The core of our framework is an evolutionary search which iteratively modifies the parent model's code $a'_{\mathrm{par}}$ to generate the child model's code $a_{\mathrm{cur}}$ to discover a high-performing optimized architecture, $a^\ast$. Here, $a_0$ is initialized with a base model $a^{\mathrm{base}}$. Specifically, $a'_{\mathrm{par}}$ is mutated into an intermediate model $a'_{\mathrm{cur}}$ with an idea $\mathrm{op}_{t}$.
Then, $a'_{\mathrm{cur}}$ is transformed by the feedback loop into the final model, $a_{\mathrm{cur}}$. Finally, $a_{\mathrm{cur}}$ is checked for validity and, if valid, $a_{\mathrm{cur}}$ is trained and evaluated. The evaluation metric (e.g., accuracy $\mathrm{acc(a_{\mathrm{cur}})}$) is added in the history database $\mathcal{D}_{\mathrm{hist}}$ with $a'_{\mathrm{cur}}$, $\mathrm{op}_{\mathrm{cur}}$, and HW budgets $\mathrm{budget_i}(a_{\mathrm{cur}}), i \in \{ 1, ..., n\}$ as follows:
\begin{equation}
    \mathcal{D}_{\mathrm{hist}} \leftarrow \mathcal{D}_{\mathrm{hist}} \cup \{ (a_{\mathrm{cur}}, \mathrm{acc}(a_{\mathrm{cur}}), \mathrm{op}_{\mathrm{cur}}, (\mathrm{budget}_1(a_{\mathrm{cur}}), \dots, \mathrm{budget}_n(a_{\mathrm{cur}}))) \mid \mathrm{is\_valid}(a_{\mathrm{cur}}) \}
\end{equation}
where $n$ denotes the number of HW budgets to be optimized. The parent models were chosen based on the top-$k$ strategy except for the Pareto-aware mutation. 

Typical evolutionary searches perform crossover and mutation operations to balance exploration and exploitation. Instead of defining crossover as an operation that takes two parent models to generate a child model, we introduce a relaxed definition. We introduce a multi-type mutation to balance exploration and exploitation.
%Therefore, in the first stage, we fairly sample a parent model and mutate it using an idea fairly sampled from $\mathcal{D}_{\mathrm{kn}}$. 
Specifically, \textit{mutation with fair idea sampling} ensures a diverse set of ideas is sampled equitably based on their attributes providing fair exploratory opportunities for mutation ideas. 
Next, \textit{Pareto-aware mutation} improves a wide range of models from the Pareto frontier according to the accuracy and HW budgets, thereby exploring models that are performant relative to their size which often be ignored in the simple top-$k$ sampling strategy. Then, \textit{LLM-driven iterative mutation} speeds up the search by focusing on and repeatedly improving high-performing models and ideas. Each type of mutation is applied sequentially. Furthermore, a feedback loop downsizes models that exceed HW constraints. These strategies allow FairNAD to effectively preserve diverse architectures. The pseudo code of FairNAD is provided in Appendix \ref{sec:F1_pseudo_code}.
%Each stage in FairNAD is explained in detail as follows. 

\subsubsection{Mutation with Fair Idea Sampling}

Diverse ideas from $\mathcal{D}_{\mathrm{kn}}$ are evaluated (Figure \ref{fig:fairNAD} (\textbf{I}).) A mutation idea, $\mathrm{op}_{\mathrm{cur}}$, is sampled from $\mathcal{D}_{\mathrm{kn}}$ based on its design attribute tag. We define ``fair idea sampling'' as the process of uniformly sampling the idea $\mathrm{op}_{\mathrm{cur}}$ according to the attributes in $\mathcal{D}_{\mathrm{attr}}$. In addition, the part of the code to be modified is specified according to the attribute tag of $\mathrm{op}_{\mathrm{cur}}$. 
%Specifically, when the ``granularity'' level's tag is ``operation'' or ``block and connectivity'', the modification is either applied to an existing module or used to design a novel module. This choice is uniformly sampled. 
Specifically, when the granularity level is tagged as ``operation'' or ``block and connectivity'', the idea  $\mathrm{op}_{\mathrm{cur}}$ is applied either to modify an existing module or to design a novel module, with this choice being uniformly sampled.
When the tag is ``network'',  $\mathrm{op}_{\mathrm{cur}}$ is prompted to ensure that the modification is applied at the network-level. This helps to enhance the diversity and novelty of the generated architectures while also ensuring that $\mathrm{op}_{\mathrm{cur}}$ is properly applied to the intended part of the model.

The sampling coverage can be formulated as follows. Suppose there are $N$ sub-categories, and sub-category $i$ is sampled with probability $p_i$. The probability that sub-category $i$ is never sampled after $n$ trials is $(1 - p_i)^n$. In contrast to unstructured sampling, our fair sampling strategy makes $p_i$ closer to $1/N$, yielding a coverage probability of $1 - (1 - 1/N)^n$ for each sub-category. The fair sampling strategy improves the coverage of minor categories and reduces the number of samples needed to explore all design categories.

\subsubsection{Pareto-Aware Mutation}

Simply applying the top-$k$ parent sampling strategy can be inefficient because it tends to overlook compact yet performant models on the Pareto frontier with low FLOPs or parameter sizes, thereby reducing the diversity of candidate models, potentially causing the search to stagnate. Thus, we introduce the Pareto-aware mutation (Figure \ref{fig:fairNAD} (\textbf{II})). This mutation employs a straightforward strategy to improve accuracy: model refinement through upscaling or hyperparameter adjustment. First, the $k$ parent models are sampled from the Pareto frontier using the method in NSGA-II \cite{deb2002fast}. Then, models below a certain budget threshold are upscaled (by increasing their width or depth), whereas models exceeding the threshold undergo hyperparameter tuning. Hyperparameters that are unrelated to width and depth, such as kernel size and stride, are adjusted. Several prompts are predefined for both upscaling and hyperparameter tuning and are uniformly sampled in each trial. 

\subsubsection{LLM-Driven Iterative Mutation}

To accelerate the search process,  an iterative refinement is introduced, which mutates high-performing ideas and models recorded in the history (Figure \ref{fig:fairNAD} (\textbf{III}).) The mutation record $h_{t-1}$ for $a'_{\mathrm{par}}$ taken from $\mathcal{D}_{\mathrm{hist}}$ and $ \mathcal{D}_{\mathrm{attr}}$ are prompted to the LLM to generate a mutation operation from a similar design attribute category. Specifically, the mutation prompt, $p_t$, is formatted with a template $\mathcal{T}_{\mathrm{idea}}$ as 
$p_t = \mathrm{format}(\mathcal{T}_{\mathrm{idea}}; a'_{\mathrm{par}}, h_{t-1}, \mathcal{D}_{\mathrm{attr}})$. This operation is repeated $d$ times.

\subsection{Feedback Loop}

For each mutation process, the generated code $a'_{\mathrm{cur}}$ is verified and modified using a feedback loop with multiple verifiers to generate the final code $a_{\mathrm{cur}}$. As a result, the chance of generating a valid code increases, which in turn improves the diversity of the generated architectures. Each verifier is described as follows: 

\begin{itemize}
%\small
\item Execution verification: $a'_{\mathrm{cur}}$ is verified to determine if it is compilable and executable. If it fails, a \textbf{debug} operation is performed; the error message and $a'_{\mathrm{cur}}$ are queried into an LLM to fix $a'_{\mathrm{cur}}$.  
\item Budget verification: The HW budgets (e.g., FLOPs and parameter size) are calculated for $a'_{\mathrm{cur}}$. If it exceeds the budgets, \textbf{model downscaling} is performed; $a'_{\mathrm{cur}}$ and the model downscaling strategy (e.g., depth or width shrinking) are provided to an LLM. Model downscaling strategies are predefined and uniformly sampled for each trial.
\item Structural verification: $a'_{\mathrm{cur}}$ is verified to ensure that the code has been modified from $a'_{\mathrm{par}}$ and retains the multi-layer architecture. If $a'_{\mathrm{cur}}$ is unmodified or consists of only a single layer, it is discarded, and the generation process is repeated with a new prompt.
\end{itemize}
%\end{tcolorbox}

\section{Experimental Setup}

$\mathcal{D}_\mathrm{kn}$, was created by scraping papers from \cite{yang2025nader}, selecting 1,149 from 2,353 abstracts to extract 15,323 ideas. For $\mathcal{D}_{\mathrm{attr}}$, 15 main categories and 55 sub-categories were generated.
$\mathcal{D}_\mathrm{ref}$ contains 14 reference architectures, including convolution-based (e.g., ResNet \cite{he2016deep}, MobileNetV2 \cite{sandler2018mobilenetv2}, MobileNetV3 \cite{howard2019searching}, EfficientNet \cite{tan2019efficientnet}, Inception \cite{szegedy2015going}, ConvNeXT \cite{liu2022convnet}, ResNeXT \cite{xie2017aggregated}), Transformer-based (e.g., Vision Transformer \cite{dosovitskiy2021an}, Swin Transformer \cite{liu2021swin}, EfficientViT \cite{cai2023efficientvit}), MLP-based (e.g., MLP-Mixer \cite{tolstikhin2021mlp}), and hybrid models (e.g., EfficientFormer \cite{li2022efficientformer}, LeViT \cite{graham2021levit}, MobileViT \cite{mehta2022mobilevit}).

FairNAD is evaluated on the widely-used NAS-Bench-201 benchmark \cite{dong2020bench}, using the CIFAR-10, CIFAR-100 \cite{krizhevsky2009learning}, and ImageNet16-120 datasets \cite{chrabaszcz2017downsampled}.
To ensure a fair comparison on NAS-Bench-201, our experimental setup follows that of prior work \cite{yang2025nader}. Specifically, the search is constrained to the maximum architecture size defined in the NAS-Bench-201 search space. The number of parameters is limited to 1.5M for all datasets, while FLOPs are constrained to 0.2 GFLOPs for CIFAR-10 and CIFAR-100, and 0.05 GFLOPs for ImageNet16-120. Although the base model can be any architecture, the results from a search starting with the CIFAR-10 version of ResNet-32 \cite{he2016deep} are reported. Results with other base models are reported in Appendix \ref{sec:B3_different_base_model}. The training process adheres to the NAS-Bench-201 settings, including its optimization settings, training epochs, and data splits. The search was conducted on four A100 GPUs with 40GB memory, using two parallel streams per GPU, resulting in eight model parallel evaluations. The experiments are implemented using the OpenMMLab library \cite{githubOpenMMLab}. Qwen3-8B \cite{yang2025qwen3} was used as the LLM. More details of the search settings and all prompts can be found in Appendix \ref{sec:A_experimental_setup_details} and \ref{sec:F2_prompts}, respectively.

\section{Experimental Results}

\subsection{Comparison with State-of-the-Art Methods}

\begin{table*}[t]
\centering
\caption{Benchmarking on NAS-Bench-201 \cite{dong2020bench}. The $\diamondsuit$ indicates methods with parameter sharing, $\heartsuit$ indicates methods without parameter sharing, and $\bigstar$ indicates methods based on LLMs. The $\pm$ denotes the s.t.d. over three runs.
}
\label{tab:nasbench}
\scalebox{0.7}{\begin{tabular}{c|l|c|ll|ll|ll}
\toprule
 & \multirow{2}{*}{Method} & Search & \multicolumn{2}{c|}{CIFAR-10} & \multicolumn{2}{c|}{CIFAR-100} & \multicolumn{2}{c}{ImageNet16-120} \\
 & &  (archs) & validation & test & validation & test & validation & test \\
  \midrule
 & ~~~ResNet \cite{he2016deep} (base model) &- & 90.83 & 93.97 & 70.42 & 70.86 & 44.53 & 43.63 \\
 & ~~~\textit{NAS-Bench-201 Optimal} &- & 91.61 & 94.37 & 73.49 & 73.51 & 46.77 & 47.31 \\ \midrule
\parbox{2.5mm}{\multirow{16}{*}{\rotatebox[origin=c]{90}{hand crafted search space}}} 
 & $\diamondsuit$ ENAS~\cite{pham2018enas} & - & 37.51$\pm$3.19 & 53.89$\pm$0.58 & 13.37$\pm$2.35 & 13.96$\pm$2.33 & 15.06$\pm$1.95 & 14.84$\pm$2.10 \\
 & $\diamondsuit$ DARTS \cite{liu2018darts}  & - & 39.77$\pm$0.00 &  54.30$\pm$0.00 &  38.57$\pm$0.00 & 15.61$\pm$0.00 &  18.87$\pm$0.00 &  16.32$\pm$0.00 \\
 & $\diamondsuit$ SETN~\cite{dong2019one} & - & 84.04$\pm$0.28 & 87.64$\pm$0.00 & 58.86$\pm$0.06 & 59.05$\pm$0.24 & 33.06$\pm$0.02 & 32.52$\pm$0.21  \\
 & $\diamondsuit$ DSNAS \cite{hu2020dsnas}  & - & 89.66$\pm$0.29 & 93.08$\pm$0.13 &  30.87$\pm$16.40 & 31.01$\pm$16.38 & 40.61$\pm$0.09 & 41.07$\pm$0.09 \\
 & $\diamondsuit$ PC-DARTS \cite{Xu2020PC} & -  & 89.96$\pm$0.15 & 93.41$\pm$0.30 & 67.12$\pm$0.39 & 67.48$\pm$0.89 & 40.83$\pm$0.08 & 41.31$\pm$0.22 \\
 & $\diamondsuit$ SNAS \cite{xie2018snas} & - & 90.10$\pm$1.04 & 92.77$\pm$0.83 & 69.69$\pm$2.39 & 69.34$\pm$1.98 & 42.84$\pm$1.79 & 43.16$\pm$2.64 \\ 
 & $\diamondsuit$ iDARTS \cite{zhang2021idarts} & - & 89.86$\pm$0.60 & 93.58$\pm$0.32 &  70.57$\pm$0.24 & 70.83$\pm$0.48 &  40.38$\pm$0.59 & 40.89$\pm$0.68 \\
 & $\diamondsuit$ GDAS \cite{dong2019searching} & - & 89.89$\pm$0.08 & 93.61$\pm$0.09 & 71.34$\pm$0.04 & 70.70$\pm$0.30 & 41.59$\pm$1.33 &  41.71$\pm$0.98  \\
 & $\diamondsuit$ DRNAS \cite{chen2021drnas} & - & {91.55$\pm$0.00} & 94.36$\pm$0.00  &  73.49$\pm$0.00 & 73.51$\pm$0.00 & 46.37$\pm$0.00 & 46.34$\pm$0.00  \\
 & $\diamondsuit$ $\beta$-DARTS \cite{ye2022b} & -  & {91.55$\pm$0.00} & 94.36$\pm$0.00  & 73.49$\pm$0.00 & 73.51$\pm$0.00 & 46.37$\pm$0.00 & 46.34$\pm$0.00  \\
 & $\diamondsuit$ $\Lambda$-DARTS \cite{movahedi2022lambda} & - & {91.55$\pm$0.00} & 94.36$\pm$0.00  &  73.49$\pm$0.00 & 73.51$\pm$0.00 & 46.37$\pm$0.00 & 46.34$\pm$0.00  \\ 
 & $\heartsuit$ REA~\cite{real2019regularized} & 500 &91.19$\pm$0.31 & 93.92$\pm$0.30& 71.81$\pm$1.12 & 71.84$\pm$0.99 & 45.15$\pm$0.89 & 45.54$\pm$1.03 \\
 & $\heartsuit$ RS~\cite{bergstra2012random} & 500 & 90.93$\pm$0.36  & 93.70$\pm$0.36& 70.93$\pm$1.09 & 71.04$\pm$1.07& 44.45$\pm$1.10 & 44.57$\pm$1.25 \\
 & $\heartsuit$ REINFORCE~\cite{williams1992simple} & 500 & 91.09$\pm$0.37 & 93.85$\pm$0.37 & 71.61$\pm$1.12 & 71.71$\pm$1.09 & 45.05$\pm$1.02 & 45.24$\pm$1.18 \\
 & $\heartsuit$ BOHB~\cite{falkner2018bohb} & 500 & 90.82$\pm$0.53 & 93.61$\pm$0.52 & 70.74$\pm$1.29 & 70.85$\pm$1.28 & 44.26$\pm$1.36 & 44.42$\pm$1.49 \\
 & $\bigstar$ GENIUS~\cite{zheng2023can} & 10 & 91.07$\pm$0.20 & 93.79$\pm$0.09 & 70.96$\pm$0.33 & 70.91$\pm$0.72 & 45.29$\pm$0.81 & 44.96$\pm$1.02 \\ 
 & $\bigstar$ CoLLM-NAS~\cite{li2025collm} & 100 & 91.59$\pm$0.04 & 94.37$\pm$0.01 & 73.44$\pm$0.12 & 73.44$\pm$0.15 & 46.62$\pm$0.10 & 46.79$\pm$0.28 \\
 % \rowcolor{blue!20}
 % & ~~~~\textit{Optimal} &- & 91.61 & 94.37 & 73.49 & 73.51 & 46.77 & 47.31 \\ \midrule
\midrule
 \parbox{2.5mm}{\multirow{6}{*}{\rotatebox[origin=c]{90}{open-ended NAS}}}
 & $\bigstar$ LLMatic~\cite{nasir2024llmatic} & 2000 & \;\; - & 94.26$\pm$0.13 & \;\; - & 71.62$\pm$1.73 & \;\; - & 45.87$\pm$0.96 \\
 & $\bigstar$ LeMo-NADe(Gemini)~\cite{rahman2024lemo} & 30 & 82.94 & 81.76 & 52.12 & 52.96 & 30.34 & 31.02 \\
 & $\bigstar$ LeMo-NADe(GPT4)~\cite{rahman2024lemo} & 30 & 90.90 & 89.41 & 63.38 & 67.90 & 27.05 & 27.70 \\
 & $\bigstar$ NADER \cite{yang2025nader} & 500 &  91.55 & 94.62 & 75.72 & 76.00 & \underline{50.20} & \underline{50.52} \\
 \cdashline{2-9} % \hdashline
 % & \multirow{3}{*}{$\bigstar$ LLMasTool} & 0 & 90.83 & 93.97 & 70.42 & 70.86 & 44.53 & 43.63 \\
 & \multirow{2}{*}{$\bigstar$ FairNAD} & 100 & \underline{91.77$\pm$0.29}& \underline{94.93$\pm$0.17}& \underline{76.54$\pm$0.28}& \underline{76.26$\pm$0.33} & 49.99$\pm$0.80 & 49.75$\pm$0.61 \\
 &  & 500 & \textbf{92.20$\pm$0.14}& \textbf{95.46$\pm$0.10}& \textbf{78.75$\pm$0.99} & \textbf{78.17$\pm$1.25} & \textbf{52.73$\pm$1.74}& \textbf{52.87$\pm$1.32}\\
 \bottomrule
 \end{tabular}}
\end{table*}

The comparisons of FairNAD with state-of-the-art NAS methods, both hand-crafted and LLM-based methods, on NAS-Bench-201 are presented in Table \ref{tab:nasbench}. The results after searching for 100 and 500 new architectures are reported. The results are presented as the mean $\pm$ standard deviation over three runs. Code-based NAS, such as GENIUS \cite{zheng2023can} and LLMatic \cite{nasir2024llmatic}, enhances flexibility by removing predefined search spaces, yet their performance is constrained by the LLM's intrinsic knowledge, limiting significant gains over hand-crafted methods. In contrast, LLM-based NAS that uses paper knowledge, like NADER \cite{yang2025nader}, surpasses the state-of-the-art (SOTA) of hand-crafted NAS by exploring broader design spaces derived from external sources like academic papers. 

%Our work advances this open-ended paradigm. 
FairNAD outperforms NADER, the current SOTA, achieving test accuracy improvements of 0.84, 2.17, and 2.35 points on CIFAR-10, CIFAR-100, and ImageNet16-120, respectively. This substantial progress can be attributed to our method's ability to structure the open-ended space into a high-quality, expansive search domain, which is then effectively and efficiently explored by our proposed FairNAD framework. The architectural design of FairNAD provided advantages for incorporating innovative modules and flexible topologies, allowing multi-level modifications to its structure (e.g., operations, connectivity) and components (e.g., convolutions). See Appendix \ref{sec:D2_discovered_architecture_details} for details. The contribution of each component is further analyzed in our ablation study.

\subsection{Ablation Studies}
\begin{table}[t]
\centering
\caption{Ablation study on CIFAR-100 dataset. The grayed components are our proposal.}
\label{tab:ablation}
\scalebox{0.85}{
\begin{tabular}{ccccc}
\hline
Planning & Mutation target & Sampling & Stages & Val / Test \\
\hline
- (Base net) & - & - & - & 70.42 / 70.86 \\
\hline
%Predefined &  & Uniform & \cellcolor{gray!30} Idea-based mut. & - \\
LLM's idea &  & Uniform & \cellcolor{gray!30} Idea-based mut. & 72.48 / 72.31 \\
Paper insp. &  & Uniform & \cellcolor{gray!30} Idea-based mut. & 74.78 / 74.55 \\
\cellcolor{gray!30} Structured insp. &  & Uniform & \cellcolor{gray!30} Idea-based mut. & 75.25 / 74.92 \\
\cellcolor{gray!30} Structured insp. & \cellcolor{gray!30} $\checkmark$ & Uniform & \cellcolor{gray!30} Idea-based mut. & 75.56 / 75.24 \\
\cellcolor{gray!30} Structured insp. & \cellcolor{gray!30} $\checkmark$ & \cellcolor{gray!30} Fair & \cellcolor{gray!30} Idea-based mut. & 77.54 / 77.05 \\
\cellcolor{gray!30} Structured insp. & \cellcolor{gray!30} $\checkmark$ & \cellcolor{gray!30} Fair & \cellcolor{gray!30} Idea-based + Pareto mut. & 77.82 / 77.28 \\
\cellcolor{gray!30} Structured insp. & \cellcolor{gray!30} $\checkmark$ & \cellcolor{gray!30} Fair & \cellcolor{gray!30} Idea-based + LLM mut. & 75.86 / 75.64 \\
\cellcolor{gray!30} Structured insp. & \cellcolor{gray!30} $\checkmark$ & \cellcolor{gray!30} Fair & \cellcolor{gray!30} Idea-based + Pareto + LLM mut. & \textbf{78.32} / \textbf{77.97} \\
\hline
\end{tabular}
}
\end{table}
\textbf{Effect of structured model design knowledge.}
FairNAD employs a structured model design knowledge attribute tree, $D_{\mathrm{attr}}$, to extract ideas from papers. This process filters out overly domain-specific ideas and enables the extraction of more fine-grained and high-quality knowledge. This innovation potentially improves the performance of the searched model. To investigate the effect of structured representation, the ablation experiments were conducted using ideas generated by different methods, as shown in Table \ref{tab:ablation}. The results after searching for 200 new architectures are reported. The first mutation stage only is conducted (denoted as ``Idea-based mut.''). Using paper-based inspiration, which employs ideas from NADER \cite{yang2025nader}, the test accuracy improves by 2.24 points compared to when LLM's ideas are used. This suggests that extracting diverse ideas from papers improves performance. Our proposed structured inspiration yields an additional 0.37-point improvement, demonstrating the effectiveness of using $D_{\mathrm{attr}}$ to curate higher-quality knowledge.

\textbf{Effect of defining mutation target.} Tagging each idea with its granularity and main and sub-category attributes using $D_{\mathrm{attr}}$ provides fine-grained control over its implementation. Specifically, the ``granularity'' tag was used to define the mutation target. This approach yields 0.32-point improvement in test accuracy, demonstrating its effectiveness in generating more diverse modules. 

\textbf{Effect of fair idea sampling.} FairNAD employs a fairness constraint in idea sampling to facilitate an unbiased and uniform exploration of the open-ended search space. The introduction of the fair sampling mechanism (denoted as ``fair'' in Table \ref{tab:ablation}) significantly improves test accuracy by 1.81 points, demonstrating the importance of broadly evaluating ideas with diverse attributes. Note that idea sampling is enabled by the proposed structured representation. Thus, the complete use of the structured representation yields a total improvement of 2.50 points over the unstructured knowledge pool. Further ablation experiments are presented in Appendix \ref{sec:B2_idea_ablation}.

\textbf{Effect of multi-type mutation.} Our proposed multi-type mutation is designed to efficiently explore the open-ended design space. The introduction of the Pareto-aware mutation stage (denoted as ``Pareto mut.'' in Table \ref{tab:ablation}) improves test accuracy by 0.23 points by facilitating the sampling of diverse parent models, which mitigates search stagnation. Further integrating the LLM-driven iterative mutation stage (denoted as ``LLM mut.'') yields an additional 0.69-point improvement. This stage accelerates the search by effectively exploring promising models and mutation operations. However, introducing the LLM-driven iterative mutation stage alone is detrimental, degrading test accuracy by 1.41 points. This is because LLM-driven iterative mutation by itself cannot generate innovative mutation ideas, making it inefficient when applied in isolation. It becomes effective when combined with the Pareto-aware mutation stage, as the latter introduces diversity in both parent models and mutation methods.

\begin{wraptable}{r}{0.45\textwidth}
    \centering
    \caption{Ablation studies of the feedback loop. The execution, budget, and structural verification improve the PR and performance.}
    \label{tab:feedback loop}
    \scalebox{0.85}{
    \begin{tabular}{lccc}
    \toprule
    \thead{Verifier} & \thead{PR} & \thead{CIFAR-100 \\ validation} & \thead{CIFAR-100 \\ test} \\
    \midrule
    \cellcolor{gray!30}All & \cellcolor{gray!30}\textbf{0.56} & \cellcolor{gray!30}\textbf{78.32} & \cellcolor{gray!30}\textbf{77.97} \\
    W/o exec. & 0.52 & 77.40 & 76.98 \\
    W/o budget & 0.04 & 77.04 & 76.63 \\
    W/o struct. & - (1.00) & 77.14 & 76.59 \\
    \bottomrule
    \end{tabular}
}
\end{wraptable}

\textbf{Effect of feedback loop.}
%The proposed feedback loop verifies and refines each model to ensure its validity. 
As shown in Table \ref{tab:feedback loop}, eliminating (denoted as ``w/o'') the execution (denoted as ``exec.''), budget, and structural (denoted as struct.) verifiers degrades the test accuracy by 0.99, 1.34, and 1.38 points, respectively. This suggests that each stage of the feedback loop contributes to generating a valid model and improves the overall performance. Specifically, eliminating the budget verifier significantly degrades the pass rate (PR) which not only impairs performance but also increases the time required to generate valid models. The additional experiments and discussion are provided in Appendix \ref{sec:C_additional_experiments} and \ref{sec:E_discussion}, respectively.

\section{Conclusions}
We introduced a novel approach for effective and efficient exploration of the open-ended design space. First, we proposed a semi-automatic method to structure the model design knowledge attributes to extract diverse and curated knowledge. Next, we introduced FairNAD, using a multi-type mutation to efficiently explore the vast open-ended design space. First, the model design ideas are uniformly sampled according to their attribute. Second, the Pareto-aware mutation stage explores compact yet performant models. Third, the LLM-driven iterative mutation accelerates the search. Furthermore, the feedback loop improves the architectural diversity by downscaling the oversized models. FairNAD achieves state-of-the-art performance, significantly outperforming existing open-ended NAS methods, and we believe it establishes a crucial foundation for future research in open-ended NAS.

%\begin{ack}
%Use unnumbered first level headings for the acknowledgments. All acknowledgments go at the end of the paper before the list of references. Moreover, you are required to declare funding (financial activities supporting the submitted work) and competing interests (related financial activities outside the submitted work). More information about this disclosure can be found at: \url{https://neurips.cc/Conferences/2026/PaperInformation/FundingDisclosure}.

%Do {\bf not} include this section in the anonymized submission, only in the final paper. You can use the \texttt{ack} environment provided in the style file to automatically hide this section in the anonymized submission.
%\end{ack}

%\section*{References}

%References follow the acknowledgments in the camera-ready paper. Use unnumbered first-level heading for the references. Any choice of citation style is acceptable as long as you are consistent. It is permissible to reduce the font size to \verb+small+ (9 point) when listing the references. Note that the Reference section does not count towards the page limit.
%\medskip

\newpage
{
\small
\bibliographystyle{abbrvnat}
\bibliography{main}
}

%{
%\small

%[1] Alexander, J.A.\ \& Mozer, M.C.\ (1995) Template-based algorithms for connectionist rule extraction. In G.\ Tesauro, D.S.\ Touretzky and T.K.\ Leen (eds.), {\it Advances in Neural Information Processing Systems 7}, pp.\ 609--616. Cambridge, MA: MIT Press.

%[2] Bower, J.M.\ \& Beeman, D.\ (1995) {\it The Book of GENESIS: Exploring Realistic Neural Models with the GEneral NEural SImulation System.}  New York: TELOS/Springer--Verlag.

%[3] Hasselmo, M.E., Schnell, E.\ \& Barkai, E.\ (1995) Dynamics of learning and recall at excitatory recurrent synapses and cholinergic modulation in rat hippocampal region CA3. {\it Journal of Neuroscience} {\bf 15}(7):5249-5262. 

%}

%%%%%%%%%%%%%%%%%%%%%%%%%%%%%%%%%%%%%%%%%%%%%%%%%%%%%%%%%%%%

\newpage
\appendix

\startcontents[appendix]
\printcontents[appendix]{l}{1}{\setcounter{tocdepth}{3}}
\newpage

\section{Experimental Setup Details} \label{sec:A_experimental_setup_details}

\subsection{NAS-Bench-201 Settings}
\begin{table*}[h]
    \centering
    \caption{NAS-Bench-201 Hyperparameter Settings}
    \label{tab:A1_nasbench201_settings}
    \begin{tabular}{ll||ll}
    \hline
    optimizer    & SGD        & initial LR    & 0.1                    \\
    Nesterov     & \checkmark & ending LR     & 0                      \\
    momentum     & 0.9        & LR schedule   & cosine                 \\
    weight decay & 0.0005     & epoch         & 200                    \\
    batch size   & 256        & normalization & \checkmark             \\
    random flip  & p=0.5      & random crop   & size=32, padding=4     \\
    \hline
    \end{tabular}
\end{table*} 
We followed the experimental setup of NAS-Bench-201 \cite{dong2020bench} for the data splits (training, validation, and test) and training hyperparameters. The hyperparameters for CIFAR-10 and CIFAR-100 are summarized in Table \ref{tab:A1_nasbench201_settings}. For ImageNet16-120, we applied the same settings, with the exception of the random crop size and padding, which were set to 16 and 2, respectively, following prior works \cite{liu2018darts, yang2025nader}.

\subsection{Evolution Parameters}
\begin{table*}[h]
    \centering
    \caption{Evolution Parameters}
    \label{tab:A2_evolution_parameters}
    \begin{tabular}{ll}
    \hline
    \multicolumn{2}{l}{\textbf{Arch. per stage}} \\
    Stage I (Mutation with fair sampling): & 8 \\
    Stage II (Pareto-aware mutation):          & 8 \\
    Stage III (LLM-driven iterative mutation):           & 4 with 2 iterative steps ($d$)\\
    \hline
    \multicolumn{2}{l}{\textbf{Thresholds}} \\
    \multicolumn{2}{l}{\parbox[t]{\dimexpr\linewidth-2\tabcolsep\relax}{To switch between scale-up and hyperparameter adjustment in stage II: \\ $0.9 \times \tau_i \quad \forall i \in \{1, \ldots, n\}$}} \\
    %\multicolumn{2}{l}{To switch between scale-up and hyperparameter adjustment in stage II: $0.9 \times \tau_i \quad \forall i \in \{1, \ldots, N\}$} \\
    \hline
    \multicolumn{2}{l}{\textbf{Max Retries}} \\
    Debug: 2                               & Model reduction: 4 \\
    \hline
    \end{tabular}
\end{table*}
The hyperparameters for the evolution search are detailed in Table \ref{tab:A2_evolution_parameters}. The number of architectures per stage (denoted as ``Arch. per stage'') is set to eight. For the model refinement stage,  the operations switch between scale-up and hyperparameter adjustment at a threshold of 90\% of the maximum value for each computation budget (e.g., denoted as $\tau_i$. We used parameter size and FLOPs.) Within the feedback loop, the debug operation (in the \textbf{execution} verification phase) is performed up to twice and the model downscaling operation (in the \textbf{budget} verification phase) is performed up to four times.

\subsection{Additional Information on the Design Knowledge Collection}
%For scraped data from a particular source (e.g., website), the copyright and terms of service of that source should be provided.
In this study, we analyzed academic papers published as open access to collect model design knowledge. The paper texts were scraped from the CVPR 2023 website (\url{https://openaccess.thecvf.com/CVPR2023}).The copyright for each paper belongs to the original author(s), and this research was conducted under Creative Commons licenses (primarily CC BY), ensuring that attribution to the authors was maintained. It should be noted that the full texts of the papers are not redistributed or published in this study; their use is strictly limited to analytical purposes.

\subsection{Experiment Compute Resources}
Evaluations were performed using four 40GB A100 GPUs. Benchmarking the search for 500 new architectures required 24.22, 26.56, and 30.06 GPU days for the CIFAR-10, CIFAR-100, and ImageNet16-120 datasets, respectively. The full research project required more than the experiments reported in this paper, including:
\begin{itemize}
    \item Preliminary experiments: For instance, we assessed the feasibility of the proposed stages of FairNAD by utilizing a reduced number of training epochs (e.g., one epoch per candidate architecture) for evaluation, in contrast to the full training schedule (e.g., 200 epochs).
    \item Exploration of alternative approaches: We also evaluated various parent sampling methods other than top-$k$ sampling. However, we found that applying Pareto sampling to all stages resulted in an inefficient search. Consequently, we concluded that Pareto sampling is best suited exclusively for the Pareto-aware mutation, while other mutations should be performed with top-$k$ sampling for faster search. Moreover, the conventional crossover method, which references two parent models, did not perform effectively with our LLM-based approach.
\end{itemize}

\section{Additional Experimental Results} \label{sec:C_additional_experiments}

\subsection{Detailed Ablation Studies of Search Strategy and Knowledge Database}

\begin{table}[h]
\centering
\caption{Detailed ablation studies of search strategy and knowledge database}
\label{tab:B1_comparison_nader}
\begin{tabular}{llcc}
\toprule
Search strategy & Knowledge DB & CIFAR-100 validation & CIFAR-100 test \\
\midrule
Multi-agent        & Unstructured & 73.58 & 73.07 \\
Multi-agent        & Structured   & 75.04 & 74.77 \\
Multi-type mutation & Unstructured & 77.14 & 77.05 \\
Multi-type mutation & Structured   & 78.32 & 77.97 \\
\bottomrule
\end{tabular}
\end{table}

To cleanly separate FairNAD's contributions, multi-type mutation and structured knowledge database, we conducted a full 2×2 controlled experiment on CIFAR-100 that disentangles the knowledge representation (unstructured vs. structured) and the search strategy (NADER multi-agent vs. our multi-type mutation) under the same LLM setup. We report results of searching for 200 new architectures in Table \ref{tab:B1_comparison_nader}. The results show that both components contribute to the performance. With the same unstructured knowledge pool, our multi-type mutation outperforms the NADER-style multi-agent search by 3.98 points in test accuracy, 77.05\% vs. 73.07\%. With the same structured knowledge pool, the improvement is 3.20 points, 77.97\% vs. 74.77\%. These results suggest that the proposed multi-type mutation is more effective than the NADER-style multi-agent search in our code-based model generation setting. We attribute this to the flexibility of multi-type mutation. The multi-agent search uses a reflector to guide the LLM based on past experience, which may constrain the search direction in a flexible code-based search space and reduce model diversity. In contrast, our method uses search history only in the LLM-driven iterative mutation, while mutation with fair sampling and Pareto-aware mutation can explore candidates without relying on history, leading to more diverse model generation.

The structured knowledge pool is also beneficial for both search strategies. It improves the multi-agent search from 73.07\% to 74.77\% (1.70 points improvement) and our multi-type mutation from 77.05\% to 77.97\% (0.92 points improvement) in test accuracy. Overall, the controlled comparison confirms that both the structured knowledge pool and the multi-type mutation independently contribute to the final performance.

\subsection{Detailed Analysis of Expert-Dependency of the Knowledge Database Taxonomy}

\begin{table}[h]
\centering
\caption{Sensitivity to removing one category from each granularity on CIFAR-100}
\label{tab:B2_sensitivity_category}
\begin{tabular}{lcc}
\toprule
Removed category / granularity & CIFAR-100 validation & CIFAR-100 test \\
\midrule
- (All categories) & 78.32 & 77.97 \\
Feature extraction operators / operation-level & 77.00 & 76.65 \\
\makecell[l]{Residual connections and multi-branch architectures / \\ block and connectivity-level} & 77.74 & 78.19 \\
Network structural patterns / network-level & 78.12 & 77.75 \\
\bottomrule
\end{tabular}
\end{table}

To quantify the expert dependency of the taxonomy, we performed a sensitivity analysis by removing one main category at a time. Specifically, the most prominent category from each granularity (e.g., operation, block and connectivity, network) was removed. We report the results on CIFAR-100 within searching for 200 new architectures in Table \ref{tab:B2_sensitivity_category}. The results show that FairNAD remains effective under all three variants, although removing any main category slightly degrades performance compared with the full taxonomy. Specifically, when the feature extraction operator is removed from operation level, test accuracy decreases by 1.32 points. For other granularity levels, test accuracy remains comparable. These results suggest that the taxonomy should include categories corresponding to important architectural components. However, the method appears relatively robust to small differences in fine-grained categorization. Thus, FairNAD is not overly dependent on any single expert-defined category.

\subsection{Sensitivity to Paper Selection}
\begin{table}[h]
\centering
\caption{Sensitivity to the paper selection.}
\label{tab:B3_paper_sensitivity}
\begin{tabular}{l c c c}
\hline
Venue-Year & Search (archs) & Val & Test \\
\hline
CVPR-2025 & 200 & 77.68 & 77.59 \\
ICCV-2025 & 200 & 77.58 & 77.84 \\
\hline
\end{tabular}
\end{table}
Table \ref{tab:B3_paper_sensitivity} presentsthe results obtained using papers from different venues and years.. The search performance remained largely consistent across different sources. Our fair idea sampling strategy encourages the search to cover diverse operators in the search space, thereby mitigating potential biases caused by trends specific to a particular venue or year.

\subsection{Detailed Ablation Studies of Idea Sampling} \label{sec:B2_idea_ablation}
\begin{table}[h]
\centering
\caption{Detailed ablation studies of idea sampling on CIFAR-100 dataset.}
\label{tab:B1_idea_ablation}
\begin{tabular}{c c c | c c}
\hline
Granularity & Main category & Sub-category & Val & Test \\
\hline
- & - & - & 75.56 & 75.24 \\
% &  &  & 75.06 & 74.64 \\
\checkmark &  &  & 76.64 & 75.96 \\
 %& \checkmark & \checkmark &  & 76.28 & 75.82 \\
 & \checkmark &  & 77.54 & 77.05 \\
 %&  & \checkmark & \checkmark &  72.00 & 71.80 \\
 &  & \checkmark & 73.68 & 73.06 \\
\hline
\end{tabular}
\end{table}
Table \ref{tab:B1_idea_ablation} shows the results of considering different attribute levels for fairness in idea sampling. We report the results after searching for 200 architectures. The checkmark ($\checkmark$) indicates that ideas are sampled uniformly at the corresponding level. A dash (-) indicates that the ideas are uniformly sampled from $\mathcal{D}_{\mathrm{kn}}$ without considering any attributes. The search is performed only with the first stage, the mutation with fair idea sampling stage.

%The test accuracy degrades by 0.60 point when only the \textit{target} attributes are considered. As detailed in Section \ref{sec:C1_details_of_attribute_tree}, more categories were found for the ``performance'' category than for the ``efficiency'' category. Subsequently, more ideas were found for the ``performance'' category. Thus, introducing the sampling according to the target may have led to an overemphasis on efficiency-focused ideas causing fewer important performance-focused ideas to be sampled.

The test accuracy improves by 0.72 and 1.81 points when considering the \textit{granularity} and \textit{main category} attributes, respectively. Since the ideas are prompted by specifying the mutation target (i.e., which part of the code is to be edited), model diversity can be improved by uniformly sampling according to the granularity attribute. Utilizing the main category yields the best performance, as it enhances the diversity of the generated models. However, considering the \textit{sub-category} significantly degrades the performance as the test accuracy drops by 2.18 points. This degradation is likely caused by overly fragmented categories. Specifically, while there were 15 main categories, there were 55 sub-categories. Using  sub-categories for sampling creates an excessive number of groups, which undermines the significance of the sampling process. Furthermore, the number of sub-categories per main category is imbalanced. Thus, considering sub-categories introduces additional biases into sampling and degrades the performance. 

\subsection{Sensitivity to Different LLMs} \label{sec:B2_different_llm}
\begin{table}[h]
\centering
\caption{Sensitivity to the choice of LLMs on CIFAR-100 dataset.}
\label{tab:B2_llm_sensitivity}
\begin{threeparttable}
\begin{tabular}{l c c c c}
\hline
Method & LLM & Search (archs) & Val & Test \\
\hline
NADER \cite{yang2025nader} & GPT4o & 500 & 75.72 & 76.00 \\
NADER \cite{yang2025nader}\tnote{$\ast$} & Qwen3-8B & 500 & 75.84 & 76.40 \\
FairNAD & Qwen2.5-Coder-7B-Instruct & 200 & 76.62 & 76.25 \\
FairNAD & Qwen2.5-Coder-32B-Instruct & 200 & 78.60 & 78.83 \\
FairNAD & Qwen3-4B & 200 & 76.62 & 76.59 \\
FairNAD & Qwen3-8B & 200 & 78.32 & 77.97 \\
FairNAD & Qwen3-32B & 200 & 78.88 & 78.27 \\
\hline
\end{tabular}
 \begin{tablenotes}
\footnotesize
\item[$\ast$] We reproduced the official implementation.
\end{tablenotes}
\end{threeparttable}
\end{table}
Table \ref{tab:B2_llm_sensitivity} shows the results of FairNAD using different sizes of Qwen2.5-Coder \cite{hui2024qwen2} and Qwen3 \cite{yang2025qwen3}. We replaced the LLM for search and used Qwen3-8B for extracting model design knowledge. Since FairNAD relies on the coding ability of LLMs, the LLM's capability to implement the intent of the prompts affects search performance. However, the older variant of Qwen, Qwen2.5-Coder-7B-Instruct, which is less performant than GPT-4o \cite{hui2024qwen2}, achieved 76.25\% test accuracy on CIFAR-100 within searching for 200 new architectures. In contrast, the prior approach, NADER \cite{yang2025nader}, reaches 76.00\% test accuracy after 500 architecture searches. Similarly, the relatively lightweight variant of Qwen3, Qwen3-4B still produces high-performing models, yielding 76.59\% test accuracy for CIFAR-100 within 200 architecture search while the prior approach, NADER \cite{yang2025nader} reaches 76.40\% test accuracy after 500 architecture searches with Qwen3-8B. This suggests that even if the LLM is not fully capable of implementing each prompt, the Pareto-aware parent sampling strategy and feedback loop provide variety in the searched models, and thus yield high-performing models. The larger variant of Qwen3, Qwen3-32B performs better than the smaller variants, yielding 78.27\% test accuracy for CIFAR-100. Also, we observed that searches using Qwen3-32B find high-performing models at an earlier stage of the search than those with smaller variants. This suggests that larger LLMs may be more capable of finding better-performing models and may also accelerates the search.

\subsection{Performance with Different Base Models} \label{sec:B3_different_base_model}
\begin{table}[ht]
\centering
\caption{Sensitivity to the base architecture on CIFAR-100 dataset.}
\label{tab:B3_base_arch_sensitivity}
\begin{tabular}{lccccc}
\hline
Architecture & Search (archs) & Params [M] & GFLOPs & Val & Test \\
\hline
NAS-Bench-201 random & 0   & 1.47 & 0.17 & 71.82 & 72.02 \\
+ FairNAD            & 200 & 1.45 & 0.19 & 76.62 & 76.59 \\
\hline
MobileNetV2-0.5 \cite{sandler2018mobilenetv2}     & 0   & 0.82 & 0.01 & 66.68 & 66.11 \\
+ FairNAD            & 200 & 1.19 & 0.16 & 77.02 & 76.54 \\
\hline
ShuffleNetV2-0.5 \cite{ma2018shufflenet}     & 0   & 0.44 & 0.01 & 66.84 & 67.71 \\
+ FairNAD            & 200 & 1.50 & 0.11 & 76.46 & 76.07 \\
\hline
RepVGG-0.375     & 0   & 1.30 & 0.09 & 71.18 & 71.09 \\
+ FairNAD            & 200 & 1.39 & 0.13 & 76.82 & 76.71 \\
\hline
\end{tabular}
\end{table}

The results of searching from different base models are shown in Table \ref{tab:B3_base_arch_sensitivity}. The search was conducted under the same HW constraints with NAS-Bench-201 \cite{dong2020bench} with the number of parameters was limited to 1.5M and FLOPs were constrained to 0.2G.``NAS-Bench-201 random'' denotes a model randomly selected from the NAS-Bench-201 search space. We used the implementation from the NADER \cite{yang2025nader} GitHub repository. For MobileNetV2 \cite{sandler2018mobilenetv2} and ShuffleNetV2 \cite{ma2018shufflenet}, the width multiplier was set to 0.5, denoted as ``-0.5''. For RepVGG \cite{ding2021repvgg}, the width multiplier was set to 0.375, denoted as ``0.375''.

FairNAD consistently performs well on different base models, achieving a test accuracy of over 76\%, which represents state-of-the-art performance. Furthermore, FairNAD is robust to base models with different modules (e.g., ResNet-based and MobileNet-based) and topologies (e.g., the multi-branch architecture in RepVGG.)

\subsection{Performance on Tiny Imagenet}

\begin{table}[h]
\centering
\caption{Results on Tiny ImageNet}
\label{tab:B6_tiny_imagenet}
\begin{tabular}{lcc}
\toprule
Model & Top-1 & Top-5 \\
\midrule
Base model (MobileNetV2) & 53.14 & 77.41 \\
+ Unstructured insp.     & 55.57 & 79.53 \\
+ FairNAD                & \textbf{58.94} & \textbf{80.13} \\
\bottomrule
\end{tabular}
\end{table}

To further evaluate scalability beyond the NAS-Bench-201 setting, FairNAD was evaluated on Tiny-ImageNet \cite{wu2017tiny} using MobileNetV2 \cite{sandler2018mobilenetv2} as the base model, with a larger budget of 4.0M parameters and 0.3 GFLOPs. Each architecture was evaluated for 30 epochs, and the model with the best validation accuracy was trained from scratch for 200 epochs. As shown in Table \ref{tab:B6_tiny_imagenet}, our method improves over the already optimized MobileNetV2 baseline by 5.80 points in top-1 accuracy, suggesting that the proposed idea sampling and multi-type mutation remain beneficial beyond the NAS-Bench-201-scale setting.

\subsection{Analysis of Wall-Clock Time and LLM Tokens}
\begin{table}[h]
\centering
\caption{LLM generation and evaluation wall-clock time for one architecture (in seconds)}
\label{tab:B7_wallclock_time}
\resizebox{\columnwidth}{!}{%
\begin{tabular}{ccccc}
\toprule
\makecell{LLM code \\ generation} & \makecell{LLM structural \\ verification} & \makecell{LLM debug \\ (max trials: 2)} & \makecell{LLM model reduction \\ (max trials: 2)} & Evaluation \\
\midrule
25 & 10 & 20 & 20 & 1005 \\
\bottomrule
\end{tabular}%
}
\end{table}

\begin{table}[h]
\centering
\caption{LLM tokens for generating one architecture}
\label{tab:B7_llm_tokens}
\begin{tabular}{lcccc}
\toprule
Category & \makecell{LLM code \\ generation} & \makecell{LLM structural \\ verification} & \makecell{LLM debug \\ (max trials: 2)} & \makecell{LLM model reduction \\ (max trials: 2)} \\
\midrule
Input  & 800  & 1200 & 2200 & 1170 \\
Output & 4000 & 1100 & 2000 & 1500 \\
\bottomrule
\end{tabular}
\end{table}

We profiled our pipeline on a four-A100 setup to obtain the wall-clock time breakdown. As shown in Table \ref{tab:B7_wallclock_time}, the evaluation stage is the dominant computational bottleneck, accounting for 90\% of the total runtime, whereas the LLM generation stages account for the remaining portion at maximum. 

From Table \ref{tab:B7_llm_tokens}, to generate 500 architectures, our method uses at most 8,740k input tokens and 12,100k output tokens, compared with NADER’s reported 5,371k input tokens and 987k output tokens.  It is important to note that Tables \ref{tab:B7_wallclock_time} and \ref{tab:B7_llm_tokens} represent an upper bound, assuming that LLM debugging and model reduction are triggered at their maximum number of trials (2 each). In practice, these stages are invoked only when necessary; for example, if debugging and model reduction are each performed once per architecture, the input-token usage is reduced to approximately 5,370k, which is comparable to NADER. The higher output-token usage in our method is mainly attributed to Qwen's thinking mode.

\section{Analysis of Model Design Attribute Structuring}

\subsection{Details of Constructed Model Design Attribute Tree} \label{sec:C1_details_of_attribute_tree}
\begin{table}[ht]
\centering
\caption{Details of model design attribute tree for performance.}
\begin{tabularx}{\linewidth}{p{0.15\linewidth} p{0.3\linewidth} X}
\hline
    Granularity & Main category & Sub-category \\
    \hline
    \multirow{3}{=}{operation level} & feature extraction operators & attention, mlp, token-mixing-mlp, grouped convolution, local convolution, convolution, input subtraction pooling, channel-mixing-mlp \\ \cline{2-3} 
     & normalization & grn, batch normalization, layer normalization \\ \cline{2-3} 
     & activation & siLU, gelu, relu, relu6, hardswish \\ 
    \hline
    \multirow{4}{=}{block and connectivity level} & input encoding & position embedding, cnn stem, patch embedding \\ \cline{2-3} 
     & residual connections and multi-branch architectures & residual connections, multi-branch architecture, learnable layer scale \\ \cline{2-3} 
     & feature fusion and aggregation & multi-scale feature fusion, element-wise addition, channel concatenation \\ \cline{2-3} 
     & adaptive feature recalibration & channel attention (SE block), multi-scale attention, multi-scale feature extraction \\ 
    \hline
    \multirow{2}{=}{network level} & network structural patterns & local-global feature extraction, multi-stage hierarchical structure, hierarchical structure, stacked transformer blocks, hierarchical channel scaling, hybrid convolution and transformer architecture \\ \cline{2-3} 
     & layer arrangement and order & pre-activation, layer normalization before attention \\ 
    \hline
    \end{tabularx}
\label{tab:C1_performance_attributes}

\end{table}
\begin{table}[ht]
\centering
\caption{Details of model design attribute tree for efficiency.}
    \begin{tabularx}{\linewidth}{p{0.15\linewidth} p{0.25\linewidth} X}
    \hline
        Granularity & Main category & Sub-category \\
        \hline
        \multirow{2}{=}{operation level} & efficient convolution & depthwise convolution, grouped convolution, depthwise separable convolution \\ \cline{2-3} 
         & efficient transformer & windowed attention, reduced key dimension attention, lightweight multi-scale linear attention, patch-based transformer \\ 
        \hline
        \multirow{2}{=}{block and connectivity level} & efficient block structures & bottleneck block, grouped convolution blocks, pooling-based blocks, convnext block, local-global block structure, inverted residual block, swin block, stochastic depth \\ \cline{2-3} 
        & dense connectivity for feature reuse & - \\ 
        \hline
        \multirow{2}{=}{network level} & network-wise scaling & hierarchical feature downsampling, compound scaling, channel width scaling \\ \cline{2-3} 
         & dynamic computation & stochastic depth \\ 
        \hline
    \end{tabularx}
\label{tab:C2_efficiency_attributes}

\end{table}

Table \ref{tab:C1_performance_attributes}
 and Table \ref{tab:C2_efficiency_attributes} show details of the model design attribute tree, $\mathcal{D}_{\mathrm{attr}}$, for performance and efficiency targets, respectively. $\mathcal{D}_{\mathrm{attr}}$ was constructed using the method explained in Section \ref{sec:4.1_semi_automated_attr_structuring}. After automatically collecting the sub-category items using LLMs, the obvious misclassifications (e.g., a main category item appears as a sub-category) and noise artifacts (e.g., the erroneous inclusion of split keys ``*main category*'') were corrected. This is a rule-based post-processing step. The main categories were checked to ensure that they appear in the sub-categories and do not contain erroneous split keys.

 Thanks to the structured attribute template, $\mathcal{T}_{\mathrm{attr}}$, fine-grained and comprehensive attributes were extracted as sub-categories. As the codes of state-of-the-art reference models ($a^{\mathrm{ref}}$) were analyzed, the attributes include representative modules of modern architectures.

\subsection{Analysis of Failure Cases} \label{sec:C2_analysis_of_failure_cases_attributes}
The major failure cases for model design attribute structuring are summarized as follows:
\begin{enumerate}
    \item Overly specific attributes: The LLM often fails to follow the instruction to extract general attributes and collects modules existing only in specific models (e.g., ``input subtraction pooling''.)
    \item Inconsistent categorization: The LLM classifies the same attribute into different categories when analyzing different reference models (e.g., ''grouped convolution'' appears in multiple categories.) 
    \item Missing attributes in specific categories: Although specific main categories exist in the manual design, no corresponding attributes exist when analyzing the reference models (e.g., no sub-categories are found for ``dense connectivity for feature reuse'' in Table \ref{tab:C2_efficiency_attributes}.)
\end{enumerate}

We attribute failures (1) and (2) primarily to the LLM's capability. Specifically, (1) is caused by the LLM's limitation in following the instructions and (2) occurs due to inconsistencies in the LLMs' output. These errors may be diminished by using more performant, larger LLMs. Failure (3) is caused by the limitations of using reference models. If specific design attributes do not exist in the set of reference models, they cannot be extracted. This issue can be addressed by adding more reference models (e.g., including models designed for different tasks or domains) or different types of reference representations such as texts. Using a hand-crafted NAS search space as a reference is another direction that could potentially be useful for extracting promising design principles. 

\section{Analysis of FairNAD}
\subsection{Analysis of Evolutionary Process}

\begin{figure*}
  \centering
  \includegraphics[width=1.0\linewidth]{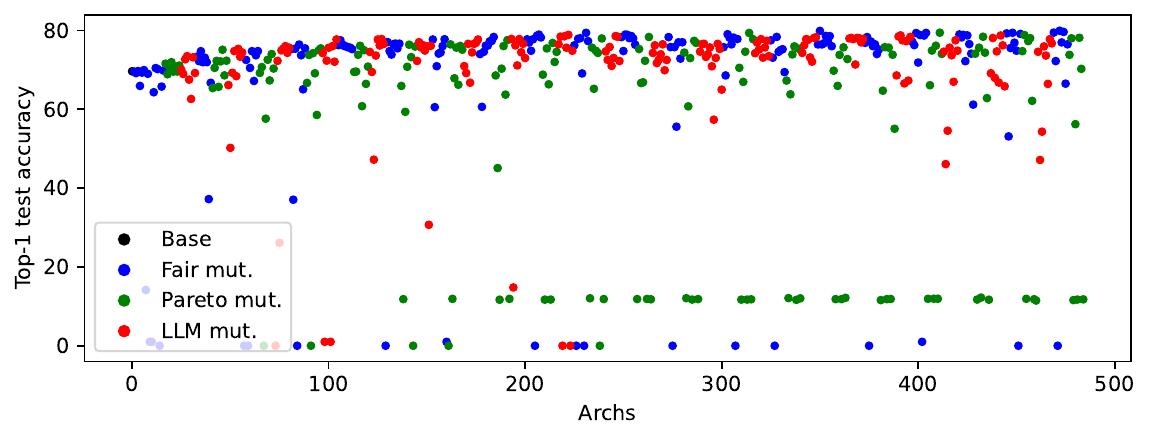}
  \caption{Evolutionary process on CIFAR-100 for searching 500 architectures.}
  \label{fig:D1_evolutionary_process}
\end{figure*}

Figure \ref{fig:D1_evolutionary_process} shows the evolutionary process on the CIFAR-100 dataset. Each mutation opertion in FairNAD contributes to improving the performance. The performance gain is significant in the first iteration of the search; performance gains can be observed in mutation with fair idea sampling (denoted as ``Fair mut.''), Pareto-aware mutation (denoted as ``Pareto mut.''), and LLM-driven iterative mutation (denoted as ``LLM mut.'').

In the early stage of search, the performance gain is larger for the LLM-driven iterative mutation than in the other two operations. This suggests that a greater focus on promising models speeds up the search when the models have not been sufficiently optimized. In the later stages of search, performance gains can be observed more often in the mutation with fair idea sampling and Pareto-aware mutation. Mutation with fair idea sampling can introduce novel modules and prevent performance from stagnating. Similarly, the Pareto-aware mutation improves diverse parent models and can potentially lead to improvements even in the later stages of search.

\subsection{Discovered Architecture Details} \label{sec:D2_discovered_architecture_details}
\definecolor{codegreen}{rgb}{0,0.6,0}
\definecolor{codegray}{rgb}{0.5,0.5,0.5}
\definecolor{codepurple}{rgb}{0.58,0,0.82}
\definecolor{backcolour}{rgb}{0.95,0.95,0.92}

\begin{lstlisting}[
language=Python,
caption={Discovered Architecture for CIFAR10 (validation accuracy: 92.18\%, test accuracy: 95.61\%, parameter size: 1.48M, FLOPs: 0.16G)},
label=D2_discovered_arch_cifar10,
backgroundcolor=\color{backcolour},   
commentstyle=\color{codegreen},
keywordstyle=\color{magenta},
numberstyle=\tiny\color{codegray},
stringstyle=\color{codepurple},
basicstyle=\ttfamily\footnotesize,
breakatwhitespace=false,         
breaklines=true,                 
captionpos=b,                    
keepspaces=true,                 
numbers=left,                    
numbersep=5pt,                  
showspaces=false,                
showstringspaces=false,
showtabs=false,                  
tabsize=2
]
import torch
import torch.nn as nn

class DPCA(nn.Module):
    def __init__(self, channel, reduction=16):  # Reduced reduction for more expressive attention
        super().__init__()
        self.avg_pool = nn.AdaptiveAvgPool2d(1)
        self.max_pool = nn.AdaptiveMaxPool2d(1)
        self.fc = nn.Sequential(
            nn.Linear(channel * 2, channel // reduction, bias=False),
            nn.ReLU(inplace=True),
            nn.Linear(channel // reduction, channel, bias=False),
            nn.Sigmoid()
        )

    def forward(self, x):
        b, c, _, _ = x.size()
        y_avg = self.avg_pool(x).view(b, c)
        y_max = self.max_pool(x).view(b, c)
        y = torch.cat([y_avg, y_max], dim=1)
        y = self.fc(y).view(b, c, 1, 1)
        return x * y

class BaseBlock(nn.Module):
    def __init__(self, in_channels, out_channels, stride=1):
        super().__init__()
        self.se_block = DPCA(out_channels)
        self.op1 = nn.Conv2d(in_channels, in_channels, kernel_size=3, stride=stride, padding=1, bias=False, groups=in_channels)
        self.op2 = nn.BatchNorm2d(in_channels)
        self.op3 = nn.PReLU(num_parameters=in_channels)  # Restored PReLU for stable gradient flow
        self.op4 = nn.Conv2d(in_channels, out_channels, kernel_size=1, stride=1, padding=0, bias=False)
        self.op5 = nn.BatchNorm2d(out_channels)
        self.downsample = (in_channels != out_channels) or (stride != 1)
        if self.downsample:
            self.proj = nn.Conv2d(in_channels, out_channels, kernel_size=1, stride=stride, padding=0, bias=False)
            self.bn = nn.BatchNorm2d(out_channels)
        self.op6 = nn.PReLU(num_parameters=out_channels)  # Restored PReLU for stable gradient flow

    def forward(self, x):
        x1 = self.op1(x)
        x1 = self.op2(x1)
        x1 = self.op3(x1)
        x1 = self.op4(x1)
        x1 = self.op5(x1)
        if self.downsample:
            x = self.proj(x)
            x = self.bn(x)
        x1 = x1 + x
        x1 = self.se_block(x1)
        x1 = self.op6(x1)
        return x1

class FeaturePool(nn.Module):
    def __init__(self):
        super().__init__()
        self.stored_features = None

    def forward(self, x):
        B, C, H, W = x.shape
        pooled = x.mean(dim=0)  # (C, H, W)
        self.stored_features = pooled
        return x

class Network(nn.Module):
    def __init__(self, num_classes=10):
        super().__init__()
        self.stem = nn.Sequential(
            nn.Conv2d(3, 64, kernel_size=3, padding=1, bias=False),
            nn.BatchNorm2d(64),
            nn.PReLU(num_parameters=64)
        )
        layers = []
        for _ in range(5):  # Original block count preserved
            layers.append(BaseBlock(64, 64, stride=1))
        self.layers1 = nn.Sequential(*layers)
        self.feature_pool = FeaturePool()  # New module for dynamic feature reuse
        layers = []
        layers.append(BaseBlock(64, 192, stride=2))  # Original channel scaling preserved
        for _ in range(7):  # Original block count preserved
            layers.append(BaseBlock(192, 192))
        self.layers2 = nn.Sequential(*layers)
        layers = []
        layers.append(BaseBlock(192, 384, stride=2))  # Original channel scaling preserved
        for _ in range(5):  # Original block count preserved
            layers.append(BaseBlock(384, 384))
        self.layers3 = nn.Sequential(*layers)
        self.avg_pool = nn.AdaptiveAvgPool2d((1, 1))
        self.fc = nn.Linear(384, num_classes)

    def forward(self, x):
        h = self.stem(x)
        h = self.layers1(h)
        h = self.feature_pool(h)  # Store batch-averaged features
        # Retrieve stored features and expand to batch size
        stored_features = self.feature_pool.stored_features
        stored_features = stored_features.unsqueeze(0).expand(x.size(0), -1, -1, -1)
        h = h + stored_features  # Add stored features to current features
        h = self.layers2(h)
        h = self.layers3(h)
        h = self.avg_pool(h)
        h = h.view(h.size(0), -1)
        h = self.fc(h)
        return h

\end{lstlisting}
\begin{lstlisting}[
language=Python,
caption={Discovered Architecture for CIFAR100 (validation accuracy: 79.90\%, test accuracy: 79.60\%, parameter size: 1.32M, FLOPs: 0.20G)},
label=D2_discovered_arch_cifar100,
backgroundcolor=\color{backcolour},   
commentstyle=\color{codegreen},
keywordstyle=\color{magenta},
numberstyle=\tiny\color{codegray},
stringstyle=\color{codepurple},
basicstyle=\ttfamily\footnotesize,
breakatwhitespace=false,         
breaklines=true,                 
captionpos=b,                    
keepspaces=true,                 
numbers=left,                    
numbersep=5pt,                  
showspaces=false,                
showstringspaces=false,
showtabs=false,                  
tabsize=2
]
import torch
import torch.nn as nn
import torch.nn.functional as F

class BandLimitedReLU(nn.Module):
    def __init__(self, c=1.0):
        super().__init__()
        self.c = nn.Parameter(torch.tensor(c), requires_grad=True)

    def forward(self, x):
        return x * (1 + torch.tanh(self.c * x)) / 2

class BinaryPrototypeModule(nn.Module):
    def __init__(self):
        super().__init__()
        self.tanh = nn.Tanh()

    def forward(self, x):
        return self.tanh(x)

class BaseBlock(nn.Module):
    def __init__(self, in_channels, out_channels, stride=1):
        super().__init__()
        self.depthwise_conv1 = nn.Conv2d(
            in_channels, in_channels, kernel_size=3, stride=stride, padding=1, groups=in_channels, bias=False
        )
        self.pointwise_conv1 = nn.Conv2d(in_channels, out_channels, kernel_size=1, bias=False)
        self.bn1 = nn.BatchNorm2d(out_channels)
        self.relu1 = BandLimitedReLU()  # Replaced with BandLimitedReLU
        
        self.depthwise_conv2 = nn.Conv2d(
            out_channels, out_channels, kernel_size=3, stride=1, padding=1, groups=out_channels, bias=False
        )
        self.pointwise_conv2 = nn.Conv2d(out_channels, out_channels, kernel_size=1, bias=False)
        self.bn2 = nn.BatchNorm2d(out_channels)
        self.relu2 = BandLimitedReLU()  # Replaced with BandLimitedReLU

        # Dual-branch attention module for gradient descent and inertial terms
        self.attention = DualBranchAttention(out_channels)

        self.downsample = (in_channels != out_channels) or (stride != 1)
        if self.downsample:
            self.proj_conv = nn.Conv2d(in_channels, out_channels, kernel_size=1, stride=stride, bias=False)
            self.proj_bn = nn.BatchNorm2d(out_channels)
        self.scale = nn.Parameter(torch.tensor(1.0))  # Learnable layer scaling parameter

    def forward(self, x):
        x1 = self.depthwise_conv1(x)
        x1 = self.pointwise_conv1(x1)
        x1 = self.bn1(x1)
        x1 = self.relu1(x1)
        
        x1 = self.depthwise_conv2(x1)
        x1 = self.pointwise_conv2(x1)
        x1 = self.bn2(x1)
        x1 = self.relu2(x1)
        x1 = self.attention(x1)
        
        if self.downsample:
            x = self.proj_conv(x)
            x = self.proj_bn(x)
        x = x * self.scale  # Apply learnable layer scaling to the residual term
        x1 = x1 + x
        return x1

class DualBranchAttention(nn.Module):
    def __init__(self, in_channels):
        super().__init__()
        # ISCA (Inertial Scale-aware Context Aggregation) branch
        self.isca_branch = nn.Sequential(
            nn.GroupNorm(16, in_channels),
            nn.AdaptiveAvgPool2d(1),
            nn.Conv2d(in_channels, in_channels // 16, kernel_size=1),
            BandLimitedReLU(),
            nn.Conv2d(in_channels // 16, in_channels, kernel_size=1),
            nn.Sigmoid()
        )
        # PGCA (Position-Guided Context Aggregation) branch
        self.pgca_branch = nn.Sequential(
            nn.Conv2d(in_channels, 1, kernel_size=1, padding=0),
            nn.Sigmoid()
        )
        # Binary Prototype Module
        self.binary_prototype = BinaryPrototypeModule()

    def forward(self, x):
        isca = self.isca_branch(x)
        pgca = self.pgca_branch(x)
        combined = isca * pgca
        x = x * combined
        x = self.binary_prototype(x)
        return x

class AdaptiveFeatureAttention(nn.Module):
    def __init__(self, in_channels):
        super().__init__()
        self.norm = nn.GroupNorm(num_groups=16, num_channels=in_channels)
        self.channel_attention = nn.Sequential(
            nn.AdaptiveAvgPool2d(1),
            nn.Conv2d(in_channels, in_channels // 16, kernel_size=1),
            BandLimitedReLU(),
            nn.Conv2d(in_channels // 16, in_channels, kernel_size=1),
            nn.Sigmoid()
        )
        self.spatial_attention = nn.Sequential(
            nn.Conv2d(in_channels, 1, kernel_size=1, padding=0),
            nn.Sigmoid()
        )

    def forward(self, x):
        x = self.norm(x)
        x = x * self.channel_attention(x)
        x = x * self.spatial_attention(x)
        return x


class Network(nn.Module):
    def __init__(self, num_classes=100):
        super().__init__()
        # Increased stem channels for scale-up
        self.stem = nn.Sequential(
            nn.Conv2d(3, 128, kernel_size=3, padding=1, bias=False),
            nn.BatchNorm2d(128),
            nn.ReLU(inplace=True)
        )
        
        # Reintroduced precompute layer with modified normalization
        self.precompute = nn.Sequential(
            nn.Conv2d(128, 128, kernel_size=1, padding=0, bias=False),
            nn.GroupNorm(16, 128),  # Changed from BatchNorm to GroupNorm
            BandLimitedReLU()
        )
        
        # Enhanced layers with adjusted depth and channel scaling
        layers1 = []
        for _ in range(2):  # Reduced from 3 to 2 blocks in layers1
            layers1.append(BaseBlock(128, 128, stride=1))
        self.layers1 = nn.Sequential(*layers1)
        self.dropout1 = nn.Dropout2d(0.1)  # Added dropout to prevent overfitting
        
        layers2 = []
        layers2.append(BaseBlock(128, 128, stride=2))  # Reduced from 256 to 128
        for _ in range(3):  # Increased from 2 to 3 additional blocks in layers2
            layers2.append(BaseBlock(128, 128))
        self.layers2 = nn.Sequential(*layers2)
        self.dropout2 = nn.Dropout2d(0.15)  # Increased dropout rate for deeper layers
        
        layers3 = []
        layers3.append(BaseBlock(128, 512, stride=2))  # Doubled from 128 to 512
        for _ in range(1):  # Reduced from 2 to 1 additional blocks in layers3
            layers3.append(BaseBlock(512, 512))
        self.layers3 = nn.Sequential(*layers3)
        self.dropout3 = nn.Dropout2d(0.2)  # Increased dropout rate for deeper layers
        
        self.avg_pool = nn.AdaptiveAvgPool2d((1, 1))
        self.fc = nn.Linear(512, num_classes)

    def forward(self, x):
        h = self.stem(x)
        h = self.precompute(h)
        h = self.layers1(h)
        h = self.dropout1(h)
        h = self.layers2(h)
        h = self.dropout2(h)
        h = self.layers3(h)
        h = self.dropout3(h)
        h = self.avg_pool(h)
        h = h.view(h.size(0), -1)
        h = self.fc(h)
        return h

\end{lstlisting}
\begin{lstlisting}[
language=Python,
caption={Discovered Architecture for ImageNet16-120 (validation accuracy: 53.93\%, test accuracy: 53.67\%, parameter size: 0.65M, FLOPs: 0.04G)},
label=D2_discovered_arch_imagenet16,
backgroundcolor=\color{backcolour},   
commentstyle=\color{codegreen},
keywordstyle=\color{magenta},
numberstyle=\tiny\color{codegray},
stringstyle=\color{codepurple},
basicstyle=\ttfamily\footnotesize,
breakatwhitespace=false,         
breaklines=true,                 
captionpos=b,                    
keepspaces=true,                 
numbers=left,                    
numbersep=5pt,                  
showspaces=false,                
showstringspaces=false,
showtabs=false,                  
tabsize=2
]
import torch
import torch.nn as nn

class PositionalEncoding(nn.Module):
    def __init__(self, d_model):
        super().__init__()
        self.d_model = d_model

    def forward(self, x):
        B, C, H, W = x.shape
        i = torch.arange(H, device=x.device).view(1, 1, H, 1)
        j = torch.arange(W, device=x.device).view(1, 1, 1, W)
        pe = torch.zeros(B, C, H, W, device=x.device)
        for c in range(C):
            freq = 10000 ** (2 * (c // 2) / self.d_model)
            pe[:, c, :, :] = torch.sin((i / freq) + (j / freq))
        return x + pe

class LocalConvAttention(nn.Module):
    def __init__(self, in_channels, out_channels, kernel_size=3, stride=1, padding=1):
        super().__init__()
        self.depthwise = nn.Conv2d(in_channels, in_channels, kernel_size, stride, padding, groups=in_channels, bias=False)
        self.pointwise = nn.Conv2d(in_channels, out_channels, kernel_size=1, bias=False)
        self.bn = nn.BatchNorm2d(out_channels)
        self.relu = nn.ReLU(inplace=True)
        self.ln = nn.LayerNorm([out_channels])
        self.attention = nn.Sequential(
            nn.AdaptiveAvgPool2d(1),
            nn.Flatten(),
            nn.Linear(out_channels, 2),
            nn.Sigmoid(),
        )

    def forward(self, x):
        x = self.depthwise(x)
        x = self.pointwise(x)
        x = self.bn(x)
        x = self.relu(x)
        x = x.permute(0, 2, 3, 1)
        x = self.ln(x)
        x = x.permute(0, 3, 1, 2)
        attn = self.attention(x)
        attn = attn.view(attn.size(0), 2, 1, 1)
        mean, _ = attn.chunk(2, dim=1)
        x = x + mean
        return x

class BaseBlock(nn.Module):
    def __init__(self, in_channels, out_channels, stride=1):
        super().__init__()
        self.local_conv_attention = LocalConvAttention(in_channels, out_channels, stride=stride)
        self.op4 = nn.Conv2d(out_channels, out_channels, kernel_size=1, stride=1, padding=0, bias=False)
        self.op5 = nn.BatchNorm2d(out_channels)
        self.downsample = (in_channels != out_channels) or (stride != 1)
        if self.downsample:
            self.proj = nn.Conv2d(in_channels, out_channels, kernel_size=1, stride=stride, padding=0, bias=False)
            self.bn = nn.BatchNorm2d(out_channels)
        self.op6 = nn.ReLU(inplace=True)

    def forward(self, x):
        x1 = self.local_conv_attention(x)
        x_concat = x1
        x_concat = self.op4(x_concat)
        x_concat = self.op5(x_concat)
        if self.downsample:
            x = self.proj(x)
            x = self.bn(x)
        x_concat = x_concat + x
        x_concat = self.op6(x_concat)
        return x_concat

class Network(nn.Module):
    def __init__(self, num_classes=120):
        super().__init__()
        self.stem = nn.Sequential(
            nn.Conv2d(3, 51, kernel_size=3, padding=1, bias=False),
            PositionalEncoding(51),
            nn.BatchNorm2d(51),
            nn.ReLU(inplace=True)
        )
        layers = []
        for _ in range(8):
            layers.append(BaseBlock(51, 51, stride=1))
        self.layers1 = nn.Sequential(*layers)
        layers = []
        layers.append(BaseBlock(51, 102, stride=2))
        for _ in range(15):
            layers.append(BaseBlock(102, 102))
        self.layers2 = nn.Sequential(*layers)
        layers = []
        layers.append(BaseBlock(102, 204, stride=2))
        layers.append(BaseBlock(204, 204, stride=2))  # Added second downsampling block
        self.layers3 = nn.Sequential(*layers)
        self.avg_pool = nn.AdaptiveAvgPool2d((1, 1))
        self.fc = nn.Linear(204, num_classes)

    def forward(self, x):
        h = self.stem(x)
        h = self.layers1(h)
        h = self.layers2(h)
        h = self.layers3(h)
        h = self.avg_pool(h)
        h = h.view(h.size(0), -1)
        h = self.fc(h)
        return h

\end{lstlisting}

In Code \ref{D2_discovered_arch_cifar10}, Code \ref{D2_discovered_arch_cifar100}, and Code \ref{D2_discovered_arch_imagenet16}, the codes for the models with the best validation accuracy, searched on CIFAR-10, CIFAR-100, and ImageNet16-120 datasets, respectively, are presented.

\textbf{The discovered architecture for CIFAR-10 (Code \ref{D2_discovered_arch_cifar10})} has the following unique characteristics:
\begin{itemize}
    \item Customized channel attention, \texttt{DPCA}: While the Squeeze-and-Excitation (SE) block, a commonly used attention mechanism, identifies important channels using only the channel-wise average values of the feature maps (\texttt{AdaptiveAvgPool2d}), the proposed DPCA leverages both average and maximum values (\texttt{AdaptiveMaxPool2d}). This combination enables the model to capture both globally and locally important features.
    \item Dynamic feature sharing mechanism, \texttt{FeaturePool}: At an intermediate point in the network (after \texttt{layers1}), the feature maps from all images in a batch are averaged, and the resulting mean feature (\texttt{stored\_features}) is temporarily stored. This mean feature is then added back to the feature map of each image in the batch. This mechanism dynamically shares global features.
    \item Deeper and wider network: The depth of \texttt{layer2} is increased (e.g., from 5 to 7) and the width is increased (e.g., from 16 to 64, 32 to 192, and 64 to 384)).
\end{itemize}

\textbf{The discovered architecture for CIFAR-100 (Code \ref{D2_discovered_arch_cifar100})} has the following unique characteristics:
\begin{itemize}
    \item Adopting depthwise separable convolution: This is technique, widely used in lightweight models like MobileNet, significantly reduces both parameters and FLOPs while maintaining or even improving the model's representational capacity.
    \item Innovative attention module, \texttt{DualBranchAttention}: By multiplying the outputs of the \texttt{isca\_branch} (e.g., channel attention) and the \texttt{pgca\_branch} (e.g., spatial attention), the model can focus on key regions within important channels. Furthermore, the \texttt{BinaryPrototypeModule} normalizes the attention-enhanced features using a tanh function, scaling their values to the range $[-1,1]$. This is expected to stabilize the inputs to subsequent layers and facilitate a smoother training process.
    \item Customized activation function \texttt{BandLimitedReLU}: This function behaves similarly to ReLU, but is smoother around the origin and utilizes a trainable parameter $c$.
    \item {Stabilizing training by layer-scaling and dropout}: Notably, as the network deepens, the dropout rate for \texttt{Dropout2d} is gradually increased (e.g., from 0.1 to 0.15 and then to 0.2.)
    \item {Shallow but wide network}: Although the depth is decreased (e.g., from 5 to 2, from 5 to 3, from 5 to 1), the width is increased (e.g., from 16 to 128, from 32 to 128, from 64 to 512).
\end{itemize}

\textbf{The discovered architecture for ImageNet16-120 (Code \ref{D2_discovered_arch_imagenet16})} has the following unique characteristics:
\begin{itemize}
    \item Using \texttt{PositionalEncoding} with a CNN-based architecture: Positional encoding, which is commonly used in Transformer models but not typically in  CNNs, is applied within the stem module.
    \item Combining multiple techniques in \texttt{LocalConvAttention} module: First, Depthwise separable convolution is used to improve computational efficiency. The mean of the attention values is then added to the entire feature map, enabling dynamic feature shifting conditioned on the input. In addition, both \texttt{BatchNorm2d} and \texttt{LayerNorm} are employed to leverage their complementary properties, thereby improving training stability.
    \item Non-standard depth and width: The architecture employs non-standard channel sizes (e.g., 51, 102, 204), differing from the common power-of-two design, along with uneven layer depths (e.g., 8, 15, 2), where the \texttt{layers2} block is notably deep. FairNAD allows for the flexible customization of channels and layers to balance performance and the computational budget.
\end{itemize}

\textbf{In summary}, the architectural design of FairNAD provides the following advantages:
\begin{itemize}
    \item It incorporates innovative modules and functions, allowing for the creation of highly flexible and customizable network topologies.
    \item It introduces modifications at multiple levels of granularity, including individual operations, connectivity, and the overall network structure.
    \item The changes are not confined to a single type of module but span diverse components, such as convolution, attention mechanisms, and pooling.
    \item Beyond individual module design, it offers the flexibility to adjust the overall width and depth of the network.
\end{itemize}

% In summary, innovative modules & arbitrary&flexible topologies, 新規(creative)な関数の設計．operation, connectivity, network-wiseの全ての粒度に変更がなされている．convolution, attention mechanismなど異なるモジュールが導入されている．モジュールの設計に留まらずwidth/depthを柔軟に変更できる

%\subsection{Attribute Choices for Fair Idea Sampling}

\subsection{Analysis of Failure Cases} \label{sec:D3_analysis_of_failure_cases_fairNAD}
A primary limitation of our method lies in the LLM's occasional failure to correctly translate a high-level idea into a functional implementation. We identified five principal failure cases:
\begin{enumerate}
    \item Incomplete generation: The LLM often truncates the output, failing to generate the complete code for complex architectures.
    \item Component hallucination: The model substitutes unknown modules or functions with plausible but non-existent or incorrect alternatives.
    \item Shape mismatch: Tensor shape mismatches frequently occur, particularly when integrating heterogeneous modules such as CNNs and Transformers.
    \item Model downscaling failure: The initially generated model becomes excessively large, causing the subsequent model downscaling step to fail.
    \item Structural verification failure: The LLM incorrectly identifies a valid model as invalid, or an invalid model as valid. Specifically, although the LLM performs well in determining whether the code has been modified, it often fails to determine whether the architecture is multi-layered.
\end{enumerate}
We attribute failures (1) and (2) primarily to the resource constraints of our experimental setup (e.g., GPU memory) and the inherent limitations of the LLM employed. These issues could potentially be mitigated by using hardware with larger memory or more capable LLMs. While our feedback loop can partially address (3) shape mismatches, (4) oversized models, and (5) structurally invalid models, we found that prompt engineering alone has limited effectiveness for these more structural problems.

As a promising direction for future work, we hypothesize that all five failure cases could be more robustly addressed by incorporating a graph-based representation, such as those proposed in NADER \cite{yang2025nader} and Genesys \cite{cheng2025language}. The graph-based representation defines the module classes or network structures. For example, Genesys \cite{cheng2025language} predefines the GPTblock, a meta module implemented in PyTorch. This module can be factorized into a tree structure of sub-modules to be explored for language models. Genesys builds a module library from external sources, and the modules in this library are then searched to serve as sub-modules. By reducing the number of code lines required for implementation, this graph-based approach lowers the chance of (1) incomplete generation. Moreover, using the module library helps prevent (2) component hallucination. Furthermore, the predefined module class potentially reduces the chance of (3) shape mismatch errors, (4) model downscaling failure, and (5) structural verification failure. The limitations of our work are further discussed in the following section.

\section{Discussion, Limitations, and Future Works} \label{sec:E_discussion}

The proposed methodology exhibits broad applicability, allowing for its integration with other NAD approaches. Firstly, our semi-automated knowledge structuring method is designed to be compatible with the idea extraction phase of NAD frameworks, making it particularly useful for sourcing design knowledge from the literature. Secondly, the distinct search stages and the feedback loop within FairNAD can be incorporated into existing NAD methods, offering a promising avenue to enhance their overall performance. Consequently, the proposed method yields positive social impacts by reducing the time and effort required for manual model design by experts and improving model performance by leveraging information from external sources such as papers to design more powerful models.

A potential negative social impact of our method is its high GPU costs for search. Like NAS methods in general, FairNAD generates, trains, and evaluates a vast number of candidate models to find the optimal architecture. This process is a major issue due to its high GPU costs and significant environmental footprint. This challenge is particularly pronounced in FairNAD, as it incrementally modifies and evaluates models, thus requiring a large number of architecture searches. Moreover, FairNAD requires a substantial number of LLM tokens for iterative improvements in the feedback loop. However, we position this cost as an investment to achieve performance levels that are unattainable with conventional methods. FairNAD demonstrates the significant potential of code-based NAS by achieving a substantial accuracy improvement; specifically, it improves test accuracy on CIFAR-100 by a remarkable 6.55\% over LLMatic, the state-of-the-art method for model exploration solely through code modification (excluding NADER, which also uses graph representations). We believe that this computational cost can be significantly reduced (e.g., it was observed that using a larger LLM, Qwen3-32B, accelerates the search) in the future through advancements in LLMs and prompt optimization, and we consider this research an important first step toward this new paradigm.

In addition to the failure cases discussed in Section \ref{sec:D3_analysis_of_failure_cases_fairNAD}, FairNAD has several limitations. First, there is a limitation on the variety of modules that can be tested. This is because FairNAD is a code-based method, and the implementable modules are constrained by the knowledge of the LLM. For a more detailed analysis, please refer to the failure case analysis in the previous section. Second, the scope of our search was limited to the backbone. In principle, FairNAD can also be applied to search for the neck and head components. However, designing the neck and head is more challenging than that of the backbone, as they are often tailored to specific tasks or domains. The current model design attribute tree is vision-specific, as it is constructed from vision models. Therefore, direct application to domains such as NLP or speech would require reconstructing the attribute tree based on representative models and literature in the target domain. Importantly, however, our method is not tied to the visual domain. The core idea is to build a domain-specific attribute tree from papers and use it to reduce bias in LLM-based inspiration sampling. Thus, the same pipeline can be applied to NLP or speech by rebuilding the attribute tree with appropriate domain-specific corpora. 

To address the aforementioned limitations, the directions for future work are as follows:
\begin{itemize}
    \item Combining FairNAD with graph-based representations: We provide a detailed explanation in Section \ref{sec:D3_analysis_of_failure_cases_fairNAD}.
    \item Reducing GPU costs for search: A coarse-to-fine approach for architectural modification could accelerate the search. For instance, larger parts of the model code could be edited in the earlier stages of the search based on the LLM's knowledge. Subsequently, smaller modifications could be applied to more promising models.
    \item Exploring open-set extension: The current pipeline relies on predefined categories; therefore, it may miss ideas that do not fit into these categories. Unmatched ideas could be retained in an ``unknown'' pool and subsequently grouped based on their semantic similarity to identify potential new sub-categories. This would allow the attribute tree to evolve as new architectural paradigms emerge.
    \item Searching neck and head architectures as well as backbones: To enable the search for more task- or domain-specific architectures, design knowledge can also be extracted from code in addition to papers. 
    \item Fine-tuning the base LLM: Fine-tuning the base LLM with previous search histories and experimental results is a feasible way to transfer knowledge across search spaces and reduce the need to start from scratch. This helps to explore larger search spaces by using the search experiences of smaller search spaces. 
\end{itemize}

\section{Implementation Details} 
\subsection{Pseudo Code} \label{sec:F1_pseudo_code}

The mutation process within FairNAD is detailed in Algorithm \ref{alg:F1_mutation}. The pseudo code of FairNAD is shown in Algorithm \ref{alg:F1_fairNAD}.
\begin{algorithm}
\caption{Mutation with a Feedback Loop}
\label{alg:F1_mutation}
\begin{algorithmic}[1]
\Require Parent model $a_{\mathrm{par}}$, Idea $\mathrm{op}_{\mathrm{cur}}$
\Ensure Generated child model $a_{\mathrm{cur}}$ or \textbf{null} on failure

\vspace{\baselineskip}
%\Procedure{MutateWithFeedback}{$a_{\mathrm{par}}, \mathrm{op}_{\mathrm{cur}}$}
\State $p_{\mathrm{cur}} \gets \text{Format}(a_{\mathrm{par}}, \mathrm{op}_{\mathrm{cur}})$
\State $a'_{\mathrm{cur}} \gets \text{LLM}(p_{\mathrm{cur}})$
\State is\_compilable $\gets$ \textbf{false}

\Comment{--- Execution Verification Phase ---}
\For{$i \gets 1 \to \text{max\_debug}$}
    \State is\_compilable, error\_msg $\gets$ CheckCompilation($a'_{\mathrm{cur}}$)
    \If{is\_compilable}
        \State \textbf{break}
    \Else
        \State $p_{\mathrm{debug}} \gets \text{Format}(a_{\mathrm{par}}, \text{error\_msg})$
        \State $a'_{\mathrm{cur}} \gets \text{LLM}(p_{\mathrm{debug}})$
    \EndIf
\EndFor
\If{\textbf{not} is\_compilable}
    \State \Return \textbf{null} \Comment{Compilation failed}
\EndIf

\Comment{--- Budget Verification Phase ---}
\State satisfies\_budget $\gets$ \textbf{false}
\For{$i \gets 1 \to \text{max\_downscale}$}
    \If{$\forall i \in \{1, \ldots, n\}, \text{budget}_i(a'_{\mathrm{cur}}) < \tau_i$}
        \State satisfies\_budget $\gets$ \textbf{true}
        \State \textbf{break}
    \Else
        \State $p_{\mathrm{downscale}} \gets \text{FormatForDownscale}(a'_{\mathrm{cur}})$
        \State $a'_{\mathrm{cur}} \gets \text{LLM}(p_{\mathrm{downscale}})$
    \EndIf
\EndFor
\If{\textbf{not} satisfies\_budget}
    \State \Return \textbf{null} \Comment{Budget constraints not met}
\EndIf

\Comment{--- Structural Verification Phase ---}
\If{IsValidStructure($a'_{\mathrm{cur}}$)}
    \State \Return $a'_{\mathrm{cur}}$ \Comment{Success}
\Else
    \State \Return \textbf{null} \Comment{Invalid structure, retry with a new idea}
\EndIf

\end{algorithmic}
\end{algorithm}

\begin{algorithm}
\caption{FairNAD}
\label{alg:F1_fairNAD}
\begin{algorithmic}[1]
\Require
    \begin{tabular}[t]{@{}l}
        Base model architecture $a^{\mathrm{base}}$;\\
        Model design knowledge database $\mathcal{D}_{\mathrm{kn}}$;
        Model design attribute tree $\mathcal{D}_{\mathrm{attr}}$; \\
        Number of initial models $k_0$;
        Number of parents for each strategy $k_1, k_2, k_3$; \\
        Number of iterative steps for LLM mutation $d$; \\
        Constraints for each budget $\tau_1, \dots, \tau_n$; \\
        Total number of generations $G$.
    \end{tabular}
\Ensure
    Optimized architecture $a^{\ast}$.

\vspace{\baselineskip}
\State \textbf{Initialization:}
\State Initialize history database $\mathcal{D}_{\mathrm{hist}} \leftarrow \emptyset$.
\State Let $a_0 \leftarrow a^{\mathrm{base}}$.
\State Sample $k_0$ ideas $\{\mathrm{op}_1, \dots, \mathrm{op}_{k_0}\}$ from $\mathcal{D}_{\mathrm{kn}}$ uniformly based on the main category attribute.
\For{$j = 1$ to $k_0$}
    \State $a_j \leftarrow \text{Mutate}(a_0, \mathrm{op}_j)$.
    \State Train and evaluate $a_j$.
    \State $\mathcal{D}_{\mathrm{hist}} \leftarrow \mathcal{D}_{\mathrm{hist}} \cup \{ (a_j, \mathrm{acc}(a_j), \mathrm{op}_{j}, (\mathrm{budget}_1(a_j), \dots, \mathrm{budget}_n(a_j)))\}$.
\EndFor

\vspace{\baselineskip}
\State \textbf{Evolutionary Loop:}
\For{generation $g = 1$ to $G$}
    \State \Comment{--- I. Mutation with fair idea sampling ---}
    %\Statex \hspace{\algorithmicindent} $\triangleright$ 1. Mutation with fair idea sampling
    \State Select $k_1$ parent models $P_1$ from $\mathcal{D}_{\mathrm{hist}}$ by using top-$k$ strategy.
    \For{each parent $a_{\mathrm{par}} \in P_1$}
        \State Sample an idea $\mathrm{op}_{\mathrm{cur}}$ from $\mathcal{D}_{\mathrm{kn}}$ uniformly based on the main category attribute.
        \State $a_{\mathrm{cur}} \leftarrow \text{Mutate}(a_{\mathrm{par}}, \mathrm{op}_{\mathrm{cur}})$.
        \State Train and evaluate $a_{\mathrm{cur}}$.
        \State $\mathcal{D}_{\mathrm{hist}} \leftarrow \mathcal{D}_{\mathrm{hist}} \cup \{ (a_{\mathrm{cur}}, \mathrm{acc}(a_{\mathrm{cur}}), \mathrm{op}_{\mathrm{cur}}, (\mathrm{budget}_1(a_{\mathrm{cur}}), \dots, \mathrm{budget}_n(a_{\mathrm{cur}})))\}$.
    \EndFor

    \vspace{\baselineskip}
    \State \Comment{--- II. Pareto-aware mutation ---}
    \State Select $k_2$ parent models $P_2$ from $\mathcal{D}_{\mathrm{hist}}$ from Pareto frontier by using the sampling method in NSGA-II.
    \For{each parent $a_{\mathrm{par}} \in P_2$}
        \If{$\exists i \in \{1, \dots, n\} \text{ s.t. } \text{budget}_i(a_{\mathrm{par}}) < 0.9 \cdot \tau_i$}
        \State $\mathrm{op}_{\mathrm{cur}} \leftarrow \text{ModelUpScalingIdea}(a_{\mathrm{par}})$
    \Else
        \State $\mathrm{op}_{\mathrm{cur}} \leftarrow \text{HyperparameterAdjustmentIdea}(a_{\mathrm{par}})$
    \EndIf
        %\State $\mathrm{op}_{\mathrm{cur}} \leftarrow \text{ModelUpScalingIdea}(a_{\mathrm{par}})$ or $\text{HyperparameterAdjustmentIdea}(a_{\mathrm{par}})$.
        \State $a_{\mathrm{cur}} \leftarrow \text{Mutate}(a_{\mathrm{par}}, \mathrm{op}_{\mathrm{cur}})$.
        \State Train and evaluate $a_{\mathrm{cur}}$.
        \State $\mathcal{D}_{\mathrm{hist}} \leftarrow \mathcal{D}_{\mathrm{hist}} \cup \{ (a_{\mathrm{cur}}, \mathrm{acc}(a_{\mathrm{cur}}), \mathrm{op}_{\mathrm{cur}}, (\mathrm{budget}_1(a_{\mathrm{cur}}), \dots, \mathrm{budget}_n(a_{\mathrm{cur}})))\}$.
    \EndFor

    \State \Comment{--- III. LLM-driven iterative mutation ---}
    \State Select $k_3$ parent models $P_3$ from $\mathcal{D}_{\mathrm{hist}}$ by using top-$k$ strategy.
    \For{each parent $a_{\mathrm{par}} \in P_3$}
        \For{$i = 1$ to $d$} \Comment{Iteratively refine the model for $d$ steps}
            \State $x \leftarrow \text{FindEntry}( \mathcal{D}_{\mathrm{hist}}, a_{\mathrm{par}})$.
            \State $\mathrm{op}_{\mathrm{cur}} \leftarrow \text{IdeaMutation}(a_{\mathrm{par}}, x, \mathcal{D}_{\mathrm{attr}})$.
            \State $a_{\mathrm{cur}} \leftarrow \text{Mutate}(a_{\mathrm{par}}, \mathrm{op}_{\mathrm{cur}})$.
            \State Train and evaluate $a_{\mathrm{cur}}$.
            \State $\mathcal{D}_{\mathrm{hist}} \leftarrow \mathcal{D}_{\mathrm{hist}} \cup \{ (a_{\mathrm{cur}}, \mathrm{acc}(a_{\mathrm{cur}}), \mathrm{op}_{\mathrm{cur}}, (\mathrm{budget}_1(a_{\mathrm{cur}}), \dots, \mathrm{budget}_n(a_{\mathrm{cur}})))\}$
            \State $a_{\mathrm{par}} \leftarrow a_{\mathrm{cur}}$. \Comment{Update the model for the next iteration of refinement}
        \EndFor
    \EndFor
\EndFor

\vspace{\baselineskip}
\State \textbf{Output:}
\State $a^{\ast} \leftarrow$ Select the architecture with the best validation accuracy from $\mathcal{D}_{\mathrm{hist}}$
\State \Return $a^{\ast}$.
\end{algorithmic}
\end{algorithm}

\newpage
\subsection{Prompts} \label{sec:F2_prompts}

%\subsubsection{Attribute Template} 
\subsubsection{Prompt for Model Design Attribute Structuring} \label{sec:F1.1_attribute_template}

\definecolor{myAttrCol}{RGB}{154, 152, 86}
\newtcolorbox{promptboxattr}[1]{
  breakable,
  colback=white,           % ボックス内の背景色
  %colframe=gray!60!black,  % 枠線の色
  colframe=myAttrCol,
  colbacktitle=myAttrCol,    % タイトル部分の背景色
  coltitle=white,          % タイトル文字の色
  %fonttitle=\bfseries,     % タイトル文字を太字に
  title=#1,                % ボックスのタイトル（引数で指定）
  boxrule=0.8pt,           % 枠線の太さ
  arc=2mm,                 % ボックスの角を丸めない（四角くする）
  top=10pt,                % ボックス内上部の余白
  bottom=10pt,             % ボックス内下部の余白
  left=10pt,               % ボックス内左側の余白
  right=10pt,              % ボックス内右側の余白
}

To structure model design attributes, the attribute template and reference models are fed to an LLM. The attribute template is used as a system prompt, while a reference model is fed to the LLM with the user prompt. 
\begin{promptboxattr}{Prompt: Attribute Template}
\textbf{Role}: You are a dedicated assistant for generating neural network model architectures' design attributes based on a given visual model.
``Model design attribute'' means specific types/categories of the key model designs.

\vspace{\baselineskip}
\textbf{Inputs}: Each time, the users will provide you with the PyTorch code of the model.

\vspace{\baselineskip}
\textbf{Goals}: Your task is to generate a list of initial model design attributes that are relevant to the given model. The generated attributes should be categorized into two lists:
\begin{enumerate}
    \item Attributes which improves performance:

    \textcolor{blue}{\{attribute\_examples\_for\_performance\_improvements\}}
    \item Attributes which improves efficiency:

    \textcolor{blue}{\{attribute\_examples\_for\_efficiency\_improvements\}}
\end{enumerate}
    Try to find attributes not in the above list as well.

\vspace{\baselineskip}
\textbf{Constraints}:
\begin{itemize}
    \item Be comprehensive
    \item Ensure that each attribute is concise, specific, and clearly describes the model's key innovations.
      For example, ``convolution'' is valid, but ``a visual module'' is too vague.
    \item Avoid duplication in performance and efficiency lists.
    \item Provide at least one attribute for each category unless it is irrelevant for the given model.
    \item The attribute must be related to the architecture design of the visual models.
    \item The attribute does NOT refer to training, optimization, initialization, dataset, pruning/quantization, pre-training or hardwares.
    \item The attribute is about visual domain. It does NOT refer to other modals such as depth, LiDAR/Radar, or point clouds.
\end{itemize}

\vspace{\baselineskip}
\textbf{Outputs}:
    Your output is two JSON objects with keys of each attribute category.

    Seperate two JSON objects with **Performance** and **Efficiency**.

    If there are no items, return an empty list.

\vspace{\baselineskip}
\textbf{Example output}

**Performance**

operation level: \{

    \hspace{2em} feature extraction operators: [convolution, self-attention, mlp],

    \hspace{2em} normalization:[layer normalization, batch normalization]

\}

block and connectivity level: \{

     \hspace{2em} input encoding: [cnn stem],

     \hspace{2em} residual connections and multi-branch architectures: [multi-branch structure]

\}

network level: \{

    \hspace{2em} network structural patterns: [hierarchical structures]

\}

\vspace{\baselineskip}
**Efficiency**

operation level: \{

    \hspace{2em} efficient convolution operation: [depthwise separable convolution]

\}

block and connectivity level: \{

    \hspace{2em} efficient block structures: [inverted residual block]

\}

network level: \{

    \hspace{2em} dynamic computation: [conditional computation]

\}

\end{promptboxattr}

\begin{promptboxattr}{Prompt: Attribute Structuring}
We provide a model code of \textcolor{blue}{\{reference\_model\_name\}}.

Considering its innovation, list-up all model attributes.

Here is the code: \textcolor{blue}{\{reference\_model\_code\}}
\end{promptboxattr}

\begin{promptboxattr}{\textcolor{blue}{Attribute Examples for Performance Improvements}}

Operation level
\begin{enumerate}
    \item Feature extraction operators:
    
    Core operations used to extract features from data. For example:
   \begin{itemize}
       \item Convolution:  Improvements such as kernel size design, dilated convolution (expanded receptive field), deformable convolution (spatially adaptive kernels), etc
       \item Self-attention: The core mechanism of Transformers. Includes multi-head attention for multi-perspective information extraction and windowed self-attention for localized computation
       \item MLP: Methods that perform feature extraction using only MLPs, without convolution or attention (e.g., MLP-Mixer)
   \end{itemize}
   
   \item Normalization:

   Normalization is essential for stabilizing and accelerating training. For example: Batch Normalization, Layer Normalization, Group Normalization, Instance Normalization.

   \item Activation:

   Nonlinearity into the network. For example: ReLU, Leaky ReLU, GeLU, Swish (SiLU)
\end{enumerate}

\vspace{\baselineskip}
Block and connectivity level
\begin{enumerate}
    \item Input encoding:
    
        Methods to encode input data. For example: CNN stem, Patch embedding, Positional encoding
    \item Residual connections and multi-branch architectures:

        Structures to enhance the diversity of feature extraction. For example: residual connections (ResNet), multi-branch structures (inception)

    \item Feature fusion and aggregation:

        Methods to combine features from different network locations (layers or branches). For example: element-wise addition, concatenation along channels (DenseNet and Inception), multi-scale feature fusion (U-Net, FPN)

    \item Adaptive feature recalibration:

        Attention mechanisms that dynamically learn which information is important. For example: channel attention (SE block), spatial attention
\end{enumerate}

\vspace{\baselineskip}
Network level
\begin{enumerate}
    \item Network structural patterns:

       High-level structure of the entire network and how the resolution of feature maps changes throughout the network. For example: hierarchical structure (Swin Transformer), feature pyramid networks (FPN)

    \item Layer arrangement and order:
    
    Order of operations from a more fine-grained perspective, such as within a module. For example: pre-activation (ResNet)
\end{enumerate}
       
\end{promptboxattr}

\begin{promptboxattr}{\textcolor{blue}{Attribute Examples for Efficiency Improvements}}
Operation level
\begin{enumerate}
    \item  Efficient convolution operation:

       Convolution operation decomposed or replaced with computationally cheaper alternatives. For example:
       depthwise separable convolution (MobileNet), grouped convolution (ResNeXt), channel shuffle (ShuffleNet)
    \item Efficient transformers:
    
       Proposes efficient self-attention mechanism, which is the main source of computational complexity in Transformers. For example, local attention (Swin Transformer), linear attention
\end{enumerate}

\vspace{\baselineskip}
Block and connectivity level
\begin{enumerate}
    \item Efficient block structures:

       Blocks that balances computational efficiency and expressive ability. For example, Inverted Residual Block (MobileNetV2), Squeeze-and-Excitation (SE) Block
    \item Dense connectivity for feature reuse:

       Features from earlier layers are aggressively reused to improve parameter efficiency. For example, Dense Connectivity (DenseNet)
\end{enumerate}

\vspace{\baselineskip}
Network level
\begin{enumerate}
    \item Network-wise scaling:

       Depth and width are scaled in a balnnced manner to improve performance efficiently. For example, Compound Scaling (EfficientNet)
    \item Dynamic computation:

       The computational path or amount of computation is dynamically adjusted, reducing average computational cost. For example, Early Exiting, Conditional Computation (Mixture-of-Experts)
\end{enumerate}

\end{promptboxattr}

\subsubsection{Prompt for Knowledge Extraction} 
First, papers related to visual backbone architectures are filtered according to their title and abstract. Then, design ideas are extracted from each paper.

\begin{promptboxattr}{Prompt: Paper Filtering by Abstract}
\textbf{Background}: You need to evaluate a given paper to determine if it can inspire to design the better basic blocks architecture of vision models.

\vspace{\baselineskip}
\textbf{Inputs}: paper title, abstract, and model attribute lists structured with level, main category and sub\-category. We give two lists for performance and efficiency improvements.

\vspace{\baselineskip}
\textbf{Goals}:
According to the title and abstract of the paper, analyzing whether it can get inspiration from the paper to design the better basic block architecture of visual models and refers to any item of the given model attribute lists.

\vspace{\baselineskip} 
\textbf{Constraints}:
\begin{itemize}
    \item The inspiration must be related to the basic block architecture design of the visual models.
    \item The idea is about model backbone, not about specific head design.
    \item The inspiration is NOT about training, optimization, initialization, dataset, hardware, or resolutions.
    \item The inspiration is useful for 2D image domain. It is NOT specialized to other modals such as depth, LiDAR/Radar, or point clouds.
    \item The inspiration falls in any category of the model attribute list.
    \item Even if the paper does not meet the above constraints, keep it if it can provide hints for model design. In this case, provide the most relevant level, main category and sub\-category.
    \item Model attribute lists are structured as:
\begin{lstlisting}
  operation level:
      main category 1: [sub-category 1, 2, ...]
      main category 2: [sub-category 1, 2, ...]
      ...
  block and connectivity level:
      main category 1: [sub-category 1, 2, ...]
      main category 2: [sub-category 1, 2, ...]
      ...
  network level:
      main category 1: [sub-category 1, 2, ...]
      main category 2: [sub-category 1, 2, ...]
      ...
\end{lstlisting}

    \item Answer with \#\#response\#\# in the end.

    \begin{itemize}
        \item If the paper refers to the model attributes, answer with the appropriate category. Split target, granularity, main category and sub\-category by ``*'' . (e.g., \#\#response\#\#target*granularity*main category*sub\_category)

        For example, \#\#response\#\#performance*operation level*feature extraction operators*attention
        \item If the paper does not refer to the model attributes, answer with no (e.g., \#\#response\#\#no)
    \end{itemize}

\end{itemize}

\vspace{\baselineskip}
\textbf{Title}: \textcolor{blue}{\{title\}}

\vspace{\baselineskip}
\textbf{Abstract}: \textcolor{blue}{\{abstract\}}

\vspace{\baselineskip}
\textbf{Design attributes for performance improvements}:

\textcolor{blue}{\{attributes\_for\_performance\_improvement\}} (Table \ref{tab:C1_performance_attributes})

\vspace{\baselineskip}
\textbf{Design attributes for efficiency improvements}:

\textcolor{blue}{\{attributes\_for\_efficiency\_improvement\}} (Table \ref{tab:C2_efficiency_attributes})

\end{promptboxattr}
\begin{promptboxattr}{Prompt: Model Design Knowledge Extraction}
You are a computer vision research expert. Get inspirations from this paper to list the ideas for designing the basic block architecture of the visual model backbone.

A paper usually contains several sections: abstract,  introduction, related work, methods, experiments and conclusion. Please focus on the methods of the paper to respond the inspirations.

\vspace{\baselineskip}
\textbf{Constraints}:
\begin{itemize}
    \item Provide ideas only related to model architecture design for visual model backbone.
    \item The idea is about model backbone, not about specific head design.
    \item Do NOT refer to training, optimization, initialization, dataset, or hardware.
    \item The idea is useful for 2D image domain. It is NOT specialized to other modals such as depth, LiDAR/Radar, or point clouds.
    \item The idea does NOT mention to use other pre-trained models.
    \item The inspiration should fall in any category of the model attribute lists. We provide two lists for attributes related to performance and efficiency improvement. The model attribute lists are structured as:

    \begin{lstlisting}
    operation level:
        main category 1: [sub-category 1, 2, ...]
        main category 2: [sub-category 1, 2, ...]
        ...
    block and connectivity level:
        main category 1: [sub-category 1, 2, ...]
        main category 2: [sub-category 1, 2, ...]
        ...
    network level:
        main category 1: [sub-category 1, 2, ...]
        main category 2: [sub-category 1, 2, ...]
        ...
    \end{lstlisting}
    
\end{itemize}

\vspace{\baselineskip}
\textbf{Output rules}:
\begin{itemize}
    \item Extract multiple ideas from the paper.
    \item Each idea must be less than 50 words. Keep it simple and exact.
    \item Do not include special characters (such as `*' or `:') in the output text.
    \item Answer the ideas with list. For example,
    \begin{itemize}
        \item idea 1
        \item idea 2
        \item ...
    \end{itemize}
    
\end{itemize}

\vspace{\baselineskip}
\textbf{Design attributes for performance improvements}:

\textcolor{blue}{\{attributes\_for\_performance\_improvement\}} (Table \ref{tab:C1_performance_attributes})

\vspace{\baselineskip}
\textbf{Design attributes for efficiency improvements}:

\textcolor{blue}{\{attributes\_for\_efficiency\_improvement\}} (Table \ref{tab:C2_efficiency_attributes})

\vspace{\baselineskip}
\textbf{Content of the paper:} \textcolor{blue}{\{paper\}}

\end{promptboxattr}

\definecolor{myFairNADCol}{RGB}{116,83,153}
\newtcolorbox{promptboxcommon}[1]{
  %enhanced,
  %attach boxed title to top left={yshift=-0.25cm, xshift=0.5cm}, % タイトルの位置を調整
  breakable,
  colback=white,           % ボックス内の背景色
  %colframe=gray!60!black,  % 枠線の色
  colframe=gray!60!,
  colbacktitle=gray!60!,    % タイトル部分の背景色
  %colframe=gray!60!,
  %colbacktitle=gray!60!,
  coltitle=white,          % タイトル文字の色
  %fonttitle=\bfseries,     % タイトル文字を太字に
  title=#1,                % ボックスのタイトル（引数で指定）
  boxrule=0.8pt,           % 枠線の太さ
  arc=2mm,                 % ボックスの角を丸めない（四角くする）
  top=10pt,                % ボックス内上部の余白
  bottom=10pt,             % ボックス内下部の余白
  left=10pt,               % ボックス内左側の余白
  right=10pt,              % ボックス内右側の余白
}

\subsubsection{Prompt for Architectural Mutation}

\textbf{Common Mutation Prompt}:

A common mutation prompt template was used for all three types of architectural mutations in FairNAD. Each idea was formatted according to its granularity tag to specify the mutation target.
\begin{promptboxcommon}{Prompt: Architectural Mutation}
\textbf{[System Prompt]}

\vspace{\baselineskip}
\textbf{Overall Instructions}

You are an expert PyTorch assistant. Your goal is to modify the given model to improve its accuracy based on the user's request.

\begin{itemize}
    \item **Plan First**: Use the Chain-of-Thought technique. Plan the modification first and then start coding.
    \item Do not change the whole block or model. The changes need to be applied to some parts of blocks (For example, `stem', `base block', `downsample block', `Network').
    \item **Respect the Request**: Faithfully implement the user's core idea.
    \item If the idea already exists, apply it to a different module or propose an improved version.
\end{itemize}

\vspace{\baselineskip}
\textbf{Technical \& output rules}
\begin{itemize}
    \item **PyTorch Only**: Do not use `tensorflow`, `keras`, or `huggingface transformers`.
    \item **Preserve Code**: Do not remove existing helper classes/functions.
    \item **Tensor Shapes**: Ensure shape compatibility. Input is \texttt{(B, 3, H, W)}, final output is \texttt{(B, num\_classes)}. Also, check the shape compatibility for intermediate features when modifications are made.
    \item **Backbone Only**: Do not add a classifier head.
    \item **mmengine**: If used, follow the registry pattern
    
    (\texttt{\string@MODELS.register\_module(force=True)}). Do not assume new \texttt{cfg} keys.
        \item **Avoid import errors**: Do not forget required imports when adding new modules. (For example, \texttt{import torch})
    \item **Final Output**:
    \begin{itemize}
        \item Provide the complete, final model code in a single block.
        \item Do not include usage examples, training code, or
        
        \texttt{if \_\_name\_\_ == `\_\_main\_\_':}.
    \end{itemize}

\end{itemize}

\vspace{\baselineskip}
\vspace{\baselineskip}
\textbf{[User Prompt]}

\vspace{\baselineskip}
\textcolor{blue}{\{idea\}} Please modify the following model: \textcolor{blue}{\{parent\_model\_code\}}

\end{promptboxcommon}

\begin{promptboxcommon}{Template: Specifying Mutation Target}
When the \textbf{idea's granularity is tagged as ``operation'' or ``block and connectivity''}, the idea is formatted with the template uniformly sampled from: 
\begin{itemize}
    \item Create and use a new, innovative module based on the idea `\textcolor{blue}{\{idea\}}'
    \item Modify any module (other than \texttt{Network}) according to the idea `\textcolor{blue}{\{insp\}}'
\end{itemize}

\vspace{\baselineskip}
When the \textbf{idea's granularity is tagged as ``network''}, the idea is formatted with the template:
\begin{itemize}
    \item Change the network structure by the idea `\textcolor{blue}{\{idea\}}'
\end{itemize}

\end{promptboxcommon}

\textbf{Ideas for Pareto-Aware Mutation}:

The ideas for Pareto-aware mutation are predefined for model upscaling and hyperparameter adjustment. The ideas are uniformly sampled when applied.

\definecolor{myParetoCol}{RGB}{60,179,113}
\newtcolorbox{promptboxpareto}[1]{
  breakable,
  colback=white,           % ボックス内の背景色
  colframe=myParetoCol!60!,
  colbacktitle=myParetoCol!60!,    % タイトル部分の背景色
  coltitle=white,          % タイトル文字の色
  title=#1,                % ボックスのタイトル（引数で指定）
  boxrule=0.8pt,           % 枠線の太さ
  arc=2mm,                 % ボックスの角を丸めない（四角くする）
  top=10pt,                % ボックス内上部の余白
  bottom=10pt,             % ボックス内下部の余白
  left=10pt,               % ボックス内左側の余白
  right=10pt,              % ボックス内右側の余白
}
\begin{promptboxpareto}{Model Upscaling Ideas}
\begin{itemize}
    \item In Network, change the number of repeats of layers in self.layers1, 2, ...to make the total number of layers larger. Set larger number at the middle of the network
    \item In Network, change the number of repeats of layers in self.layers1, 2, ...to make the total number of layers larger. Set larger number at the end of the network
    \item In Network, rewrite channel numbers to larger values while keeping the current ratios
    \item In Network, rewrite channel numbers to larger values by changing the current ratios
\end{itemize}
\end{promptboxpareto}

\begin{promptboxpareto}{Hyperparameter Adjustment Ideas}
\begin{itemize}
    \item In any block, change kernel size
    \item In any block, change stride
    \item In any block, change use\_se
    \item In any block, change activation
\end{itemize}
\end{promptboxpareto}

\subsubsection{Prompt for Idea Mutation}

In LLM-driven iterative mutation, the idea was mutated using an LLM with the template, $\mathcal{T}_{\mathrm{idea}}$ as shown in the following. When the accuracy improved in the parent model $a_{t-1}$ from the grand-parent model $a_{t-2}$, ``\texttt{Prompt: For Improved Parents}'' was used. When the accuracy dropped, ``\texttt{Prompt: For Degraded Parents}'' was used. ``Model refinement ideas'' refers to the ideas used in the Pareto-aware mutation.
\definecolor{myIdeaCol}{RGB}{165,42,42}
\newtcolorbox{promptboxidea}[1]{
  breakable,
  colback=white,           % ボックス内の背景色
  colframe=myIdeaCol!60!,
  colbacktitle=myIdeaCol!60!,    % タイトル部分の背景色
  coltitle=white,          % タイトル文字の色
  title=#1,                % ボックスのタイトル（引数で指定）
  boxrule=0.8pt,           % 枠線の太さ
  arc=2mm,                 % ボックスの角を丸めない（四角くする）
  top=10pt,                % ボックス内上部の余白
  bottom=10pt,             % ボックス内下部の余白
  left=10pt,               % ボックス内左側の余白
  right=10pt,              % ボックス内右側の余白
}
\begin{promptboxidea}{Prompt: \textcolor{blue}{Common Template} for Idea Mutation}

A parent model has been mutated using the idea related to the mutation attribute categories or simple scale-up ideas.

\vspace{\baselineskip}
Mutation attribute categories are organized as follows:
    \begin{lstlisting}
    operation level:
        main category 1: [sub-category 1, 2, ...]
        main category 2: [sub-category 1, 2, ...]
        ...
    block and connectivity level:
        main category 1: [sub-category 1, 2, ...]
        main category 2: [sub-category 1, 2, ...]
        ...
    network level:
        main category 1: [sub-category 1, 2, ...]
        main category 2: [sub-category 1, 2, ...]
        ...
    \end{lstlisting}

\vspace{\baselineskip}
Model refinement ideas are given by a list.

\vspace{\baselineskip}
\textbf{Design attributes}:

\textcolor{blue}{\{attributes\_for\_performance\_improvement\}}, \hspace{1em} 
\textcolor{blue}{\{attributes\_for\_efficiency\_improvement\}} 

\vspace{\baselineskip}
\textbf{Model refinement ideas}: \{refinement\_ideas\}
\end{promptboxidea}

\begin{promptboxidea}{Prompt: For Improved Parents}
\textcolor{blue}{\{common\_template\}}

\vspace{\baselineskip}
Changing from the parent model to the new model with has improved the performance, which shows that the mutation is crucial. Analyze the difference between the parent and mutated models for:

\vspace{\baselineskip}
\begin{itemize}
    \item Is the mutation about any of mutation attribute categories or simple scale-up ideas?
    \item Why this mutation improves performance?
\end{itemize}

\vspace{\baselineskip}
Based on the analysis, please make the mutated model even better by:
\begin{itemize}
    \item  If the mutation is about any of design attribute categories, come up with ideas that falls in the similar category of mutation attributes. 
    \item If the mutation is about model refinement, try other ideas about model scale-up.
    \item Provide the complete, final model code in a single block.
\end{itemize}

\vspace{\baselineskip}
Here is the parent model and the mutated model.

\vspace{\baselineskip}
\textbf{Parent model}: \textcolor{blue}{\{parent\_model\_code\}}

\vspace{\baselineskip}
\textbf{Mutated model}: \textcolor{blue}{\{current\_model\_code\}}
\end{promptboxidea}

\begin{promptboxidea}{Prompt: For Degraded Parents}
\textcolor{blue}{\{common\_template\}}

\vspace{\baselineskip}
Changing from the parent model to the new model with degraded performance. Analyze the difference between the parent and mutated models for:

\begin{itemize}
    \item Is the mutation about any of design attribute categories or model refinement?
    \item Why this mutation degraded the performance?
\end{itemize}

\vspace{\baselineskip}
Based on the analysis, please generate a new model based on the **parent model** that address the issue identified by:

\begin{itemize}
    \item If the mutation is about model refinement, think ideas other than model refinement.
    \item If the mutation is about any of design attribute categories, come up with ideas with the same category but opposite effects or ideas that falls in the different category.
    \item The change must NOT be a simple revert.
    \item Provide the complete, final model code in a single block.
\end{itemize}

\vspace{\baselineskip}
Here is the parent model and the mutated model.

\vspace{\baselineskip}
\textbf{Parent model}: \textcolor{blue}{\{parent\_model\_code\}}

\vspace{\baselineskip}
\textbf{Mutated model}: \textcolor{blue}{\{current\_model\_code\}}
\end{promptboxidea}

\subsubsection{Prompt for Feedback Loop}
For each mutation process, the generated code is verified and modified using a feedback loop. The prompts for debug (in the execution verification stage), model downscaling (in the budget verification stage), and structural verification are shown as follows.
\begin{promptboxcommon}{Prompt: Debug}

\textbf{[System Template]}

\vspace{\baselineskip}
You are an expert AI assistant specializing in debugging PyTorch code. Your task is to fix bugs in a given PyTorch model.

\vspace{\baselineskip}
The user will provide you with:
\begin{itemize}
    \item The PyTorch code of the failed model and its parent model.
    \item The full error message and trace-back.
    \item The original prompt that was used to generate the failed model.
\end{itemize}

\vspace{\baselineskip}
\textbf{Your core task}:

Your primary goal is to fix the bug in the provided model code while preserving all intended functionalities described in the original prompt. To achieve this, think step-by-step through the following process:

\begin{enumerate}
    \item Analyze the Error: Carefully examine the error message and trace-back to pinpoint the exact line and the root cause of the error.
    \item Understand the Intent: Read the original prompt to fully understand the intended functionality of the entire model and the specific role of the problematic module.
    \item Formulate a Fix: Based on your analysis, devise a correction strategy that addresses the root cause of the error without sacrificing any features.
    \item Implement the Correction: Rewrite the code to implement the fix, ensuring the entire model is complete and correct.
\end{enumerate}
    
**IMPORTANT**: You must NOT remove, disable, or bypass any modules or code blocks in the target model. For instance, do not remove any lines in the forward method of \textcolor{blue}{\{base\_module\_classes\}} classes.

\vspace{\baselineskip}
\textbf{Error cases}:

\vspace{\baselineskip}
**\texttt{Imports} error**
\begin{itemize}
    \item Do not forget importing `torch'
\end{itemize}

\vspace{\baselineskip}
**\texttt{CUDA out of memory} error**
\begin{itemize}
    \item The modules that were added to or modified from the parent model may be used repeatedly within the model. Please reduce the number of repeats of these modules**.
\end{itemize}

\vspace{\baselineskip}
** \texttt{NoneType} error**
\begin{itemize}
    \item A \texttt{NoneType} error often indicates that a function or module expected to return a tensor is returning None. This is frequently caused by an incomplete forward method or a missing class name for the final model. Ensure the model's implementation is complete and explicitly returns a tensor from the forward method.
\end{itemize}

\vspace{\baselineskip}
**\texttt{embed\_dim must be divisible by num\_heads} error**
\begin{itemize}
    \item Most likely to occur in the attention module such that the \texttt{embed\_dim} must be divisible by num\_heads:
\begin{lstlisting}
multihead_attn = nn.MultiheadAttention(embed_dim, num_heads)
\end{lstlisting}
      Change \texttt{num\_heads} value so that \texttt{embed\_dim // num\_heads}
\end{itemize}

\vspace{\baselineskip}
**\texttt{Size mismatch} error**
\begin{itemize}
    \item Do not change the output dimension size
\end{itemize}
    
\vspace{\baselineskip}
**\texttt{Expected more than 1 value per channel when training, got input size torch.Size([B, C, 1, 1])} error**
\begin{itemize}
    \item The resolution of features is too small. Consider making smaller strides, or other methods to increase the resolution of features.
\end{itemize}
    
\vspace{\baselineskip}
**Rules for when LayerNorm is implemented to CNN-based modules**
\begin{itemize}
    \item LayerNorm module **MUST** accept a 4D tensor `(B, C, H, W)' and output a tensor of the **exact same shape**. For example:
\begin{lstlisting}
self.layer_norm = nn.LayerNorm(embed_dim)
x = x.permute(0, 2, 3, 1)  # (N,C,H,W) -> (N,H,W,C)
x = self.layer_norm(x)
x = x.permute(0, 3, 1, 2)  # (N,H,W,C) -> (N,C,H,W)
\end{lstlisting}
\end{itemize}
    
\vspace{\baselineskip}
**Rules for when Conv1d are implemented with 4D inputs**
\begin{itemize}
    \item Conv1d must accept 4D tensor 4D tensor `(B, C, H, W)` and output a tensor of the **exact same shape**. For example:

\begin{lstlisting}
self.conv1d = nn.Conv1d(in_channels, out_channels, kernel_size)
B, C, H, W = x.shape
x = x.reshape(B, C, H * W)
x = self.conv1d(x)
x = x.reshape(B, C, H, W)
\end{lstlisting}
\end{itemize}
    
\vspace{\baselineskip}
**Rules for when Transformer or attention blocks are implemented to CNN-based modules**
\begin{itemize}
    \item CNN-Transformer Hybrid module **MUST** accept a 4D tensor `(B, C, H, W)' and output a tensor of the **exact same shape**. For example:
\begin{lstlisting}
**Example 1: Using nn.MultiheadAttention**
qkv = nn.Linear(embed_dim, embed_dim * 3)
multihead_attn = nn.MultiheadAttention(embed_dim, num_heads)  # embed_dim must be divisible by num_heads
# forward
B, C, H, W = x.shape
x = x.reshape(B, C, H*W)
x = x.permute(2, 0, 1)  # (H*W, B, C)
qkv = self.qkv(x)
q, k, v = qkv.chunk(3, dim=2)
x, _ = multihead_attn(q, k, v)
x = x.permute(1, 2, 0)  # (B, C, H*W)
x = x.reshape(B, -1, H, W)  # (B, C, H, W)

**Example 2: With LayerNorm**
# With LayerNorm, use LayerNorm before reshaping back to 4D tensor.
layer_norm = nn.LayerNorm(embed_dim)
# forward
x = x.reshape(B, C, H*W)
x = x.permute(2, 0, 1)  # (H*W, B, C)
qkv = self.qkv(x)
q, k, v = qkv.chunk(3, dim=2)
x, _ = multihead_attn(q, k, v)
x = layer_norm(x)  # (H*W, B, C)
x = x.permute(1, 2, 0)  # (B, C, H*W)
x = x.reshape(B, -1, H, W)  # (B, C, H, W)

**Example 3: Defining qkv**
qkv = nn.Linear(embed_dim, embed_dim * 3)
# forward
x = x.reshape(B, C, H*W)
x = x.permute(2, 0, 1)  # (H*W, B, C)
qkv = self.qkv(x)
q, k, v = qkv.chunk(3, dim=2)
q = q * self.scale
attn = (q @ k.transpose(-2, -1)) * self.scale
attn = attn.softmax(dim=-1)
x = attn @ v
x = x.permute(1, 2, 0)  # (B, C, H*W)
x = x.reshape(B, C, H, W)
\end{lstlisting}
\end{itemize}

\vspace{\baselineskip}
\textbf{Output rules}:
\begin{itemize}
    \item Provide the complete, corrected, and runnable model code as the final response.
    \item Do not add any explanations, comments, or apologies before the code block.
    \item Ensure all necessary imports are included at the beginning of the script.
    \item Wrap the final, complete code inside python . Do not wrap any other text or code snippets.
\end{itemize}
    
\vspace{\baselineskip}
\textbf{[User Template]}

\vspace{\baselineskip}
The target model was mutated from the parent model with the prompt \textcolor{blue}{\{prompt\}}.

This caused the error with the error message: \textcolor{blue}{\{error\}}.

Please fix the error of the target model.

\vspace{\baselineskip}
Target model: \textcolor{blue}{\{sample\_model\_code\}}

\vspace{\baselineskip}
Parent model: \textcolor{blue}{\{parent\_model\_code\}}
\end{promptboxcommon}

\begin{promptboxcommon}{Prompt: Model Downscaling}
\textbf{[System Template]}
You are an expert coding assistant. Your task is to fix a given PyTorch model to reduce FLOPs and parameter size.

\vspace{\baselineskip}
The user will provide you with:
\begin{itemize}
    \item The PyTorch code of the target model and the parent model.
    \item The prompt that was used to mutate the target model from the parent model.
\end{itemize}

\textcolor{blue}{\{model\_downscaling\_operation\_template\}}

\vspace{\baselineskip}
\textbf{Output rules}:
\begin{itemize}
    \item Do not return the parent or target model back.
    \item Provide the complete, corrected, and runnable model code as the final response.
    \item Do not add any explanations, comments, or apologies before the code block.
    \item Ensure all necessary imports are included at the beginning of the script.
    \item Wrap the final, complete code inside python. Do not wrap any other text or code snippets.
\end{itemize}

\vspace{\baselineskip}
\textbf{[User template]}

The target model was mutated from the parent model with the prompt `\textcolor{blue}{\{prompt\}}'.
Please reduce the FLOPs and parameter size of the model by `\textcolor{blue}{\{model\_dowmscaling\_idea\}}'.

\vspace{\baselineskip}
Target model: \textcolor{blue}{\{sample\_model\_code\}}

\vspace{\baselineskip}
Parent model: \textcolor{blue}{\{parent\_model\_code\}}

\end{promptboxcommon}

Model downscaling operations were selected based on the HW budgets ($\tau_i, \hspace{1em} \forall i \in \{1, \ldots, N\}$) of the mutated model. If any of the model's resulting HW costs is larger than $1.5 \times \tau_i$, the newly added module by the idea was restricted to a single use once within the network. Otherwise, depth or width shrinking is performed.

\begin{promptboxcommon}{\textcolor{blue}{Model Downscaling Operation Template}}
\textbf{[Restricting the Newly Added Module to a Single Use]}
Please reduce the FLOPs and parameter size by using the added module only once.

Here is the example:
\begin{lstlisting}
self.stem = stem(in_channels, out_channels)
layers = []
for _ in range(n):
    layers.append(block(in_channels, out_channels))
...
self.new_layer = AddedBlock(in_channels, out_channels)  # move the added module in the end of backbone
self.avg_pool = nn.AdaptiveAvgPool2d((1, 1))
self.fc = nn.Linear(out_channels, num_classes)
\end{lstlisting}

\vspace{\baselineskip}
\textbf{[Depth or Width Shrinking]}

Please reduce the FLOPs and parameter size without removing the added/modified module.
    
\end{promptboxcommon}

\begin{promptboxcommon}{\textcolor{blue}{Model Downscaling Ideas}}
\textbf{[Restricting the Newly Added Module to a Single Use]}
\begin{itemize}
    \item Ensure that the added module is used only once within the model
\end{itemize}

\vspace{\baselineskip}
\textbf{[Depth or Width Shrinking]}
\begin{itemize}
    \item  In Network, change the number of repeats of layers in self.layers1, 2, ... to make the total number of layers slightly smaller. Set larger number at the middle of the network
    \item In Network, change the number of repeats of layers in self.layers1, 2, ... to make the total number of layers slightly smaller. Set larger number at the end of the network
    \item In Network, rewrite channel numbers to slightly smaller values while keeping the current ratios
    \item In Network, rewrite channel numbers to slightly smaller values by changing the current ratios
\end{itemize}
\end{promptboxcommon}

\begin{promptboxcommon}{Prompt: Structural Verification}
The target model was mutated from the parent model.

Judge following and answer if the target model satisfies all of them:
\begin{itemize}
    \item The model architecture has been changed from the parent model. It's not enough for the code to change (including comments, outputs shapes); the built model architecture itself must be different.
    \item In Network class, the base block (or something equivalent) is repeated more than or equal to two times.
\end{itemize}
    
\vspace{\baselineskip}
Answer with \#\#response\#\# in the end. (e.g., **response**yes)

\vspace{\baselineskip}
Target model: \textcolor{blue}{\{sample\_model\_code\}}

\vspace{\baselineskip}
Parent model: \textcolor{blue}{\{parent\_model\_code\}}
\end{promptboxcommon}

%\appendix

%\section{Technical appendices and supplementary material}
%Technical appendices with additional results, figures, graphs, and proofs may be submitted with the paper submission before the full submission deadline (see above). You can upload a ZIP file for videos or code, but do not upload a separate PDF file for the appendix. There is no page limit for the technical appendices. 

%Note: Think of the appendix as ``optional reading'' for reviewers. The paper must be able to stand alone without the appendix; for example, adding critical experiments that support the main claims to an appendix is inappropriate. 

%%%%%%%%%%%%%%%%%%%%%%%%%%%%%%%%%%%%%%%%%%%%%%%%%%%%%%%%%%%%

%\newpage
%\input{checklist.tex}

\end{document}